%% file: main_arxiv.tex
\documentclass[12pt]{article}
\usepackage{amsmath}
\usepackage{amsthm}
\usepackage{graphicx}
\usepackage{enumerate}
\usepackage[numbers,super,sort&compress]{natbib}
\usepackage{url} 
\usepackage{mathrsfs}  
\usepackage{bm} 
\usepackage{lmodern}
\allowdisplaybreaks 
\usepackage{mathtools}
\usepackage{amsfonts}
\usepackage{color}
\usepackage{float}
\usepackage{array}
\usepackage{booktabs}
\usepackage{tabularx}
\usepackage{adjustbox}
\usepackage{hyperref}
\usepackage{subcaption}
\usepackage{amsfonts}
\usepackage{xr}
\usepackage{multirow}
\newtheorem{theorem}{Theorem}

\newtheorem{remark}[theorem]{Remark}
\newtheorem{corollary}{Corollary}[theorem]
\newtheorem{prop}{Proposition}

\newcommand{\blind}{1}

\include{GrandMacros}
\usepackage{xcolor}
\def\blue{\textcolor{black}}

\begin{document}

\def\spacingset#1{\renewcommand{\baselinestretch}%
{#1}\small\normalsize} \spacingset{1}


\if1\blind
{
  \title{\bf {\blue{Another look at statistical inference with \\ machine learning-imputed data}}}
  \author{Jessica Gronsbell$^{1}$\\
    and \\
    Jianhui Gao$^{1}$ \\
    and \\
    Zachary R. McCaw$^{2}$\\
    and\\
    Yaqi Shi$^{1}$\\
    and \\
    David Cheng$^{3}$\\
    $^{1}$Department of Statistical Sciences, University of Toronto, Toronto, ON\\
$^{2}$Department of Biostatistics, UNC Chapel Hill, Chapel Hill, NC\\
$^{3}$Biostatistics Center, Massachusetts General Hospital, Boston, MA}
  \maketitle
} \fi

\if0\blind
{
  \bigskip
  \bigskip
  \bigskip
  \begin{center}
    {\LARGE\bf Another look at inference after prediction}
\end{center}
  \medskip
} \fi

\hrule
\bigskip
\noindent
\textbf{Correspondence to:}\\
David Cheng \\
Postal address: 399 Revolution Drive, Suite 1068, Somerville, MA 02145\\
Email: \url{dcheng@mgh.harvard.edu}.\\
Telephone number: 617-643-4131\\

\noindent
Jessica Gronsbell \\
Postal address: 700 University Ave, Toronto, ON, Canada, M5G 1Z5\\
Email: \url{j.gronsbell@utoronto.ca}.\\
Telephone number: 416-978-3452\\
\hrule
\bigskip

\noindent%
{\it Keywords:}  Machine learning, measurement error, missing data, prediction-based inference, semi-parametric inference

\newpage
\bigskip
\begin{abstract}
\noindent
From structural biology to epidemiology, predictions from machine learning (ML) models increasingly complement costly gold-standard data, enabling faster, more affordable, and scalable scientific inquiry. In response, prediction-based (PB) inference has emerged to support statistical analysis that combines a large volume of predicted data with a small amount of gold-standard data. The goals of PB inference are twofold: (i) to mitigate bias arising from prediction error and (ii) to improve efficiency relative to classical inference based solely on gold-standard data. While early PB inference methods primarily focused on bias mitigation, improving efficiency remains an active area of research. \blue{Motivated by connections between PB inference and longstanding problems in statistics and related fields, we draw on the two-phase sampling literature to introduce an approach for Z-estimation with ML-imputed outcomes that is guaranteed to match or exceed the efficiency of classical inference, regardless of prediction quality. We demonstrate the utility of our approach through theoretical and numerical analyses as well as an application to UK Biobank data. We further establish new connections between existing PB inference approaches and foundational and contemporary statistical methods.}
\end{abstract}

\newpage

\spacingset{1.9} 
\section{Introduction}
\begin{figure}[ht]
    \centering
    \fbox{\includegraphics[scale = 0.40]{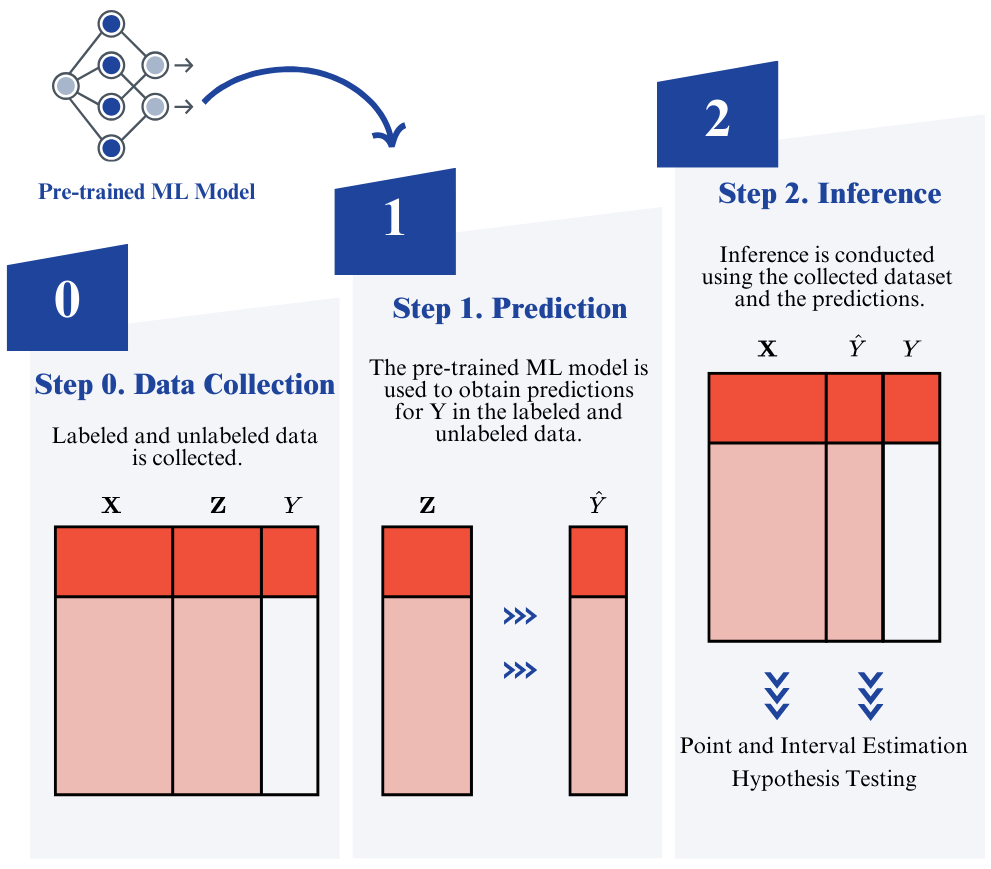}}
    \caption{\textbf{Prediction-based (PB) inference.} $Y$ is the outcome of interest, which is only partially observed, $\bZ$ is a set of auxiliary predictors of $Y$, $\Yhat$ is the prediction of $Y$ based on $\bZ$, and $\bX$ is a set of covariates. The focus of scientific inquiry is on the association between $Y$ and $\bX$. Classical statistical inference relies solely on $Y$ and $\bX$ while PB inference incorporates $\Yhat$ to make use of all of the available data. Dark red represents fully labeled observations, for which all of covariates, predictors, and outcomes are known. Light red and white represent partially labeled observations, for which covariates and predictors are known, but outcomes are missing.}
    \label{fig:figure_1}
\end{figure}

Researchers across \blue{scientific domains} are increasingly turning to machine learning (ML) models to obtain predictions for outcomes that are prohibitively time-consuming or expensive to measure directly, and then conducting statistical analyses which treat the predictions as observed data.\citep{hoffman2024we, fan2024narratives, gao2024semi, miao2024task}  This {\it{prediction-based}} (PB) inference approach, introduced by Wang et al,\cite{wang2020methods} is comprised of the two steps outlined in Figure~\ref{fig:figure_1}. First, the outcome of interest is imputed with predictions from a pre-trained model based on auxiliary data. Then, the predictions are used together with a small amount of gold-standard data to draw inferences about the parameters of interest, such as the association parameters between the outcome and a set of covariates.\footnote{PB inference has primarily focused on outcomes that are difficult to measure. Recent work has considered predicted covariates (e.g., see Kluger et al.\cite{kluger2025prediction} \blue{and Xu et al.\cite{xu2025unified}).}} In contrast to {\it{classical inference}} based solely on gold-standard data, \blue{PB inference leverages information from the predictions to enhance the efficiency and thereby the scope of scientific analyses.} 

In the last several years, applications of PB inference have arisen in diverse fields, spanning genetics, biology, and medicine. For example, population biobanks, such as the UK Biobank ($n \approx$ 500K individuals), collect high-throughput genomic data on large populations and offer unprecedented opportunities for genetic discovery. However, many phenotypes are only partially observed due to the time and expense of measurement. To enable downstream association studies with phenotypes of interest, ML models are increasingly used to impute missing outcomes based on fully observed baseline data.\citep{tong2020augmented, alipanahi2021glaucoma, cosentino2023copd, dahl2023phenotype, an2023autocomplete, mccaw2024synthetic, liu2024semiparametric, miao2024popgwas} Similarly, in structural biology, AlphaFold provides protein conformation predictions without the need for expensive and labor-intensive experiments. \citep{jumper2021highly, gan2022activation} In the analysis of electronic health record data, large language models can rapidly ascertain disease status at an accuracy competitive with manual chart review by trained annotators, facilitating clinical and epidemiological studies. \citep{yang2022large, zhou2022cancerbert, alsentzer2023zero} \\

\blue{With the growing use of ML-imputed data, researchers have proposed numerous PB inference methods to effectively analyze datasets with partially missing outcomes. \cite{wang2020methods, angelopoulos2023prediction, angelopoulos2023ppi++, miao2023assumption, gan2023prediction, egami2023using} Statistically, these methods aim to mitigate bias attributable to prediction errors and to improve efficiency relative to classical inference. One of the earliest and most influential proposals is the prediction-powered inference (\texttt{PPI}) approach.\cite{angelopoulos2023prediction} \texttt{PPI} has been applied in diverse domains\cite{miao2024task, boyeau2024autoeval, poulet2025prediction, gligoric2025can} as it does not impose any assumptions about the ML model. In particular, \texttt{PPI} provides consistent estimation and yields hypothesis tests with controlled type I error and confidence intervals with nominal coverage rates, regardless of prediction quality. This property is a key strength of \texttt{PPI} and has spurred numerous methodological extensions.\cite{demirel2024prediction, hofer2024bayesian, miao2024task} However, it has also been observed that \texttt{PPI} can have higher variance than classical inference with low quality predictions, which has led to modifications of the approach aimed at improving efficiency. \cite{angelopoulos2023ppi++, miao2023assumption, gan2023prediction}}

\blue{Similarly motivated by this observation, we revisit \texttt{PPI} and its extensions from a semiparametric efficiency perspective and introduce an alternative approach for PB inference. Our proposal provides valid inference and is also guaranteed to match or exceed the performance of classical estimators, regardless of prediction quality. Specifically, our contributions are as follows. First, we show that the \texttt{PPI} estimator for general Z-estimation problems is asymptotically linear, with an influence function analogous to that of the augmented inverse probability weighting (AIPW) estimator \citep{robins1994estimation} for a particular choice of the augmenting function. This result allows us to precisely characterize when \texttt{PPI} gains efficiency over classical estimation. Second, we adapt an approach from the two-phase sampling literature proposed by Chen \& Chen \cite{chen2000unified} (\texttt{CC}) for the purposes of PB inference and show that it provides provable efficiency gains over classical estimation. Third, we identify sufficient conditions for when the \texttt{CC} estimator gains efficiency over recent extensions of \texttt{PPI} \citep{angelopoulos2023ppi++, gan2023prediction, miao2023assumption} when estimating regression parameters in generalized linear models (GLMs) and validate this finding with comprehensive numerical studies. Fourth, we revisit the statistics and economics literature dating back to the 1960s to highlight work directly relevant to PB inference. Together, these contributions link classical approaches to this contemporary challenge to inform future methods and applications.}

We begin by formally introducing the PB inference problem and \blue{reviewing the \texttt{PPI} approach} in Section \ref{problem-setup}. We then study the asymptotic properties of the \texttt{PPI} estimator in Section \ref{robins-connection}. In Section \ref{more-eff-estimator}, we introduce the \texttt{CC} estimator and provide theoretical and conceptual comparisons with \texttt{PPI} and its extensions. In Section \ref{other-connection}, \blue{we provide new connections between the \texttt{CC} approach and modern and historical methods.} We then evaluate our findings through extensive simulation studies in Section \ref{empirical-studies}. In Section \ref{real-data}, \blue{we demonstrate the utility of the \texttt{CC} estimator in an analysis studying the relationship between android fat mass with demographic covariates among the South Asian population in the UK Biobank.} We close with a discussion in Section \ref{discussion}. 

\section{Problem set-up}\label{problem-setup}
\subsection{Notation \& parameter of interest}
\blue{We let $Y$ denote a scalar outcome and $\bX$ a $p$-dimensional vector of covariates. We focus on Z-estimation and aim to conduct statistical inference on the $q$-dimensional parameter $\bbeta^*$ defined as the unique solution to the estimating equation:
\begin{equation*}\label{param-interest-beta}
\E[ \bphi(Y, \bX; \bbeta) ] = \bm{0}
\end{equation*} 
where $\bphi(\cdot)$ is a deterministic function that satisfies conditions C.1-C.7 in the Appendix. The Z-estimation framework is broad and encompasses maximum likelihood estimation, generalized estimating equations, and the generalized method of moments, among other common estimation approaches.}

\blue{To conduct inference on $\bbeta^*$, we assume that we observe a dataset in which $\bX$ is fully observed on all observations and $Y$ is partially observed among a subset of observations indicated by $R=1$. In addition to data on $Y$ and $\bX$, we assume that predictions $\Yhat = \fhat(\bZ)$ are available for all observations based on a pre-trained ML model $\fhat$ and fully observed predictors $\bZ$. The predictors $\bZ$ may or may not include $\bX$. The assumption that $\fhat$ is pre-trained implies that its training data are independent of the data used to conduct inference on $\bbeta^*$. Consequently, $\fhat$ is treated as ``fixed'' and variation arising from its estimation is not considered in downstream inference.\footnote{Approaches that account for uncertainty in the ML model have recently been explored.\cite{gan2023prediction, zrnic2024cross}} The available data therefore consists of $n$ independent and identically distributed (iid) observations $(R_i, Y_iR_i, \bZ_i\trans, \bX_i \trans)\trans$. We denote the labeling fraction as $\pi_n = n^{-1}\sum_{i=1}^n R_i$. For simplicity and to facilitate comparisons with recent work in PB inference, we assume that the outcomes are missing completely at random (MCAR) so that $\P(R=1 \mid Y, \bX, \bZ ) = \P(R=1) = \pi$ for some $\pi\in (0,1)$.}

{\remark{\blue{With a few exceptions,\cite{chen2025unified, kluger2025prediction} existing PB inference methods\cite{angelopoulos2023prediction, angelopoulos2023ppi++, miao2023assumption, gan2023prediction} assume that the available data consist of a labeled dataset with $\nlab$ observations of $(Y_i, \bZ_i\trans, \bX_i\trans)\trans$ and an independent unlabeled dataset with $\nunlab$ observations of $(\Yhat_i, \bX_i\trans)\trans$ where $n = \nunlab + \nlab$ and $\nlab / \nunlab \to \rho$ for $\rho \in (0,1)$ as $\nunlab, \nlab \to \infty$. This set-up corresponds to a sampling scheme in which a fixed number $\nlab$
observations are selected for labeling and an additional $\nunlab$ observations are observed without labels. The labeling indicators in this set-up would therefore be constrained such that their sum is exactly $\nlab$. Our set-up assumes the observations are labeled by an independent stochastic mechanism.}}}

\subsection{Motivation for PB inference}\label{pb-motivation}
To provide general motivation for PB inference, recall that the classical estimator of $\bbeta^*$, denoted as $\bbetahatlab$, uses only the labeled observations for estimation and is the solution to:
\begin{equation*}
   n^{-1} \sum_{i = 1}^{n} \bphi(Y_i, \bX_i; \bbeta)\frac{R_i}{\pi_n} = \bm{0}. 
\end{equation*}
While $\bbetahatlab$ is consistent for $\bbeta^*$ and enables valid inference, it has limited precision when the number of labeled observations is not large. In contrast, the predictions $\Yhat$ are available in the full data, leading to a much larger dataset for analysis. However, relying exclusively on $\Yhat$ for inference can result in biased estimation depending on the quality of predictions from $\fhat$. In so-called {\emph{naive}} inference,\cite{salerno2025moment} $\Yhat$ is directly substituted for $Y$ in downstream analysis.\citep{wang2020methods, motwani2023revisiting} \blue{The parameter $\bgamma^*$ targeted by naive inference is thus the unique solution to:
\begin{equation*}\label{param-interest-beta}
\E[ \bphi(\Yhat, \bX; \bgamma)] = \bm{0}.
\end{equation*} 
The corresponding estimator based on {\blue{the full data}}, denoted as $\bgammahatall$, is the solution to:
\begin{equation*}
n^{-1} \sum_{i = 1}^{n} \bphi(\Yhat_i, \bX_i; \bgamma)  = \boldsymbol{0}.
\end{equation*}}
As the distribution of $(\Yhat, \bX\trans)\trans$ may not coincide with that of $(Y, \bX\trans)\trans$, the relationship between $\bbeta^*$  and $\bgamma^*$ is unclear, except under some restrictive situations. {\blue{For example, for least squares estimation, $\bgamma^*$ coincides with $\bbeta^*$ when $\bX$ is included in $\bZ$ and the ML model perfectly captures the true regression (i.e., the true $\E(Y \mid \bZ = {\bm{z}})$ is within the space of functions $\fhat({\bm{z}})$ can represent)}}. \blue{The application of naive inference in scientific practice therefore has the potential to produce biased, though often very precise inferences, about $\bbeta^*$. This can lead researchers to be more confident in the wrong answers, thereby undermining the scientific rationale for using predicted outcomes in statistical inference.} 
  
\subsection{Existing PB inference methods}\label{motwani-recap}
\blue{To overcome the significant limitations of naive inference, numerous PB inference methods have been proposed.}\cite{angelopoulos2023prediction, angelopoulos2023ppi++, miao2023assumption, miao2024task, mccaw2024synthetic, gan2023prediction} These methods leverage the full dataset and make use of both $Y$ and $\Yhat$ in conducting inference on $\bbeta^*$. For example, the first PB inference method, coined \texttt{PostPI}, \cite{wang2020methods} performs inference in two steps. First, the labeled data are used to model the relationship between $Y$ and $\Yhat$ to address potential inaccuracies in the predictions. Next, this model is combined with the unlabeled data to perform inference on $\bbeta^*$ using either analytic or bootstrap methods. While an innovative approach that set the stage for subsequent methodological developments, Motwani and Witten \cite{motwani2023revisiting} recently showed that \texttt{PostPI} provides valid inference only under stringent assumptions, such as when the ML model captures the true regression function.

\blue{Several years following the introduction of \texttt{PostPi}, Angelopoulos et al\citep{angelopoulos2023prediction} proposed Prediction-Powered Inference (\texttt{PPI}).} Unlike its predecessor, \texttt{PPI} has the crucial property that inference remains valid regardless of the quality of the predictions. That is, $\hat{f}$ need not {\blue{capture}} the true regression function for the estimator to remain consistent for $\bbeta^*$, for hypothesis tests to asymptotically control the type I error rate, or for confidence intervals to achieve the nominal coverage level. Formally, the \texttt{PPI} estimator of $\bbeta^*$ is the solution to the de-biased estimating equation:
\begin{align}\label{eq: ppi}
n^{-1} \sum_{i = 1}^{n} \bphi(\Yhat_i, \bX_i; \bbeta) \frac{1-R_i}{1 -\pi_n}  - \left[ \bphi(\Yhat_i, \bX_i; \bbeta) - \bphi(Y_i, \bX_i; \bbeta) \right] \frac{R_i}{\pi_n} = \bzero.
\end{align}
The trailing term of (\ref{eq: ppi}) is referred to by the authors as the ``rectifier'' and serves the purpose of correcting for potential bias in naive estimation. For the remainder of this paper, we consider a modified version of the \texttt{PPI} estimator, which we refer to as $\bbetahatppi$, that solves an estimating equation in which the leading term in (\ref{eq: ppi}) is replaced with the average over the full data, $n^{-1}\sum_{i=1}^n \bphi(\Yhat_i, \bX_i;\bbeta)$. This modification facilitates analytic comparisons with other estimators considered in this paper.

\blue{While \texttt{PPI} provides valid inference for $\bbeta^*$, three separate studies simultaneously recognized that $\bbetahatppi$ can be less efficient than $\bbetahatlab$ and proposed to incorporate data-adaptive weights within the \texttt{PPI} estimation equation to address this limitation. \cite{angelopoulos2023ppi++, miao2023assumption, gan2023prediction} Formally, these proposals are solutions to:
\begin{align}\label{eq: ppi-ext}
n^{-1} \sum_{i = 1}^{n} \bphi(Y_i, \bX_i; \bbeta) \frac{R_i}{\pi_n}  -  \widehat\bW \left[n^{-1} \sum_{i = 1}^{n}  \bphi(\Yhat_i, \bX_i; \bbeta) \frac{R_i}{\pi_n} -  n^{-1} \sum_{i = 1}^{n}  \bphi(\Yhat_i, \bX_i; \bbeta) \right] = \bzero,
\end{align}
where $\widehat\bW$ is a $q \times q$ weight matrix that controls the contribution of $\Yhat$ into estimation of  $\bbeta^*$. The first proposal, called \texttt{PPI++},\cite{angelopoulos2023ppi++} utilizes a scalar weight to minimize the trace of the resulting estimator’s asymptotic covariance matrix.\footnote{Computing confidence intervals for the \texttt{PPI} estimator is generally intractable, which is an issue that is also addressed by \texttt{PPI++}. See the supplementary materials of Angelopoulos et al.\cite{angelopoulos2023prediction} for detailed algorithms.} The second proposal, called PoSt-Prediction Adaptive (\texttt{PSPA}) inference,~\cite{miao2023assumption} utilizes a diagonal weight matrix to minimize element-wise variance. The third proposal, called Prediction De-Correlated (\texttt{PDC}) inference,\cite{gan2023prediction} utilizes a full $q \times q$ matrix to attain the lowest variance among all estimators that are solutions to (\ref{eq: ppi-ext}).\cite{xu2025unified} While \texttt{PPI++}, \texttt{PSPA}, and \texttt{PDC} achieve provable improvements over classical inference, estimators outside of the \texttt{PPI} framework (i.e., estimators that are not solutions to (\ref{eq: ppi-ext})) remain unexplored. We consider such an alternative estimator in Section \ref{more-eff-estimator}. }

\section{Asymptotic analysis of the \texttt{PPI} estimator}\label{robins-connection}
\blue{To motivate our proposal, we further analyze the \texttt{PPI} estimator and show that it can be viewed as an AIPW estimator.} AIPW estimators are asymptotically linear estimators with influence functions of the form\citep{robins1994estimation}:
\begin{align}\label{eq: aipw-if}
    {\bf{IF}}(Y, \Yhat, \bX; \bomega) =   \bA \left\{ \bphi(Y, \bX; \bbeta^*) \frac{R}{\pi} +   \bomega( \Yhat, \bX) \left(1 - \frac{R}{\pi} \right)\right\},
\end{align}
where $\bomega( \Yhat, \bX)$ is a square-integrable augmenting function of $\Yhat$ and $\bX$ and $\bA =  -\E\left[ \frac{\partial \bphi(Y, \bX; \bbeta^*)}{\partial \bbeta^{\top}} \right]^{-1}$. An estimator whose influence function has the augmenting function:
$$\bomega^{\rm opt}(\hat Y, \bX) = \E\left[ \bphi(Y, \bX; \bbeta^*) \mid \hat Y, \bX \right]$$
achieves the semiparametric efficiency bound \citep{bickel1993efficient}. \blue{In Proposition \ref{prop: betappi}, we show that the influence function of $\bbetahatppi$ takes the form of (\ref{eq: aipw-if}) for a specific choice of the augmenting function. This result formally establishes a similar finding in Angelopoulos et al\citep{angelopoulos2023prediction} for mean estimation to the broader Z-estimation problem. The proof of this and other theoretical results are provided in the Appendix for completeness.}

\begin{prop}[Asymptotic linearity of $\bbetahatppi$]\label{prop: betappi}
Under basic assumptions in the Appendix,  $\bbetahatppi$ is consistent such that $||\bbetahatppi-\bbeta^*|| = o_p(1)$. Moreover, it is asymptotically linear such that:
$$ n^{1/2} \left(\bbetahatppi - \bbeta^*\right) =  n^{-1/2}  \sum_{i=1}^n  {\bf{IF}}^{\ppi}(Y_i, \Yhat_i,  \bX_i; \bbeta^*)   + o_p(1),$$ 
where ${\bf{IF}}^{\ppi}(Y_i, \Yhat_i, \bX_i; \bbeta^*) =   \bA\left[\bphi(Y_i, \bX_i; \bbeta^*) \frac{R_i}{\pi} +  \left\{\bphi(\Yhat_i, \bX_i; \bbeta^*)-\E \left[\bphi(\Yhat_i,\bX_i;\bbeta^*)\right]\right\}(1-\frac{R_i}{\pi})  \right]$. 
\end{prop}

\blue{$\bbetahatppi$ can thus be viewed as an AIPW estimator with $\bomega^\ppi(\Yhat, \bX) = \bphi(\Yhat, \bX; \bbeta^*) - \E \left[\bphi(\Yhat,\bX;\bbeta^*)\right]$. It can be shown that this choice closely approximates $\bomega^{\rm opt}(\hat Y, \bX)$ in specific cases, such as when estimating the coefficients of a linear regression model such that $\bphi(y,\bx;\bbeta) = \bx(y-\bx\trans\bbeta)$, $\bX$ is included in $\bZ$, and the machine learning model $\hat f$ captures the true regression function. However, in general, Proposition \ref{prop: betappi} shows that the \texttt{PPI} estimator is not the most efficient estimator among AIPW estimators. Moreover, several studies\citep{angelopoulos2023ppi++, miao2023assumption, gan2023prediction} have identified that \texttt{PPI} can even be outperformed by classical inference when prediction quality is low and/or the unlabeled sample size is insufficient. In Corollary \ref{coro: betappi}, we apply Proposition \ref{prop: betappi} to identify necessary and sufficient conditions for $\bbetahatppi$ to gain efficiency over $\bbetahatlab$.}

\blue{\begin{corollary}[Efficiency of $\bbetahatlab$ vs. $\bbetahatppi$]\label{coro: betappi}
Let $\bD_{12} =  \E \left[ \bphi(Y, \bX; \bbeta^*) \bphi(\Yhat, \bX; \bbeta^*)\trans \right]$ and $\bD_{22} =  Var \left[ \bphi(\Yhat, \bX; \bbeta^*)\right]$. The difference in the asymptotic variance between
$\bbetahatlab$ and $\bbetahatppi$ is $\Delta_{\var}( \bbetahatlab, \bbetahatppi) =  (\pi^{-1} - 1) \bA \left(  2  \bD_{12} -  \bD_{22}  \right) \bA\trans$. Consequently, $\bbetahatppi$ is more efficient than $\bbetahatlab$ if and only if $2\bD_{12} -\bD_{22} \succ \bzero$.
\end{corollary}}

\blue{\remark{AIPW estimators are known to be model doubly robust in that they are $n^{1/2}$-consistent and asymptotically normal when either of two underlying nuisance parameters, typically a propensity score and a mean outcome function, are consistent at sufficiently fast rates \citep{smucler2019unifying}. For \texttt{PPI}, the mean outcome function is $\bphi(\Yhat,\bX;\bbeta^*)$ and the propensity score function is effectively an estimate of the labeling probability $\pi_n$, which is $n^{1/2}$-consistent under MCAR. Due to model double robustness, $\bbetahatppi$ remains $n^{1/2}$-consistent and asymptotically normal. However, as with other AIPW estimators, there is a contribution, $\bA\E \left[\bphi(\Yhat_i,\bX_i;\bbeta^*)\right](1-\frac{R_i}{\pi})$, to the influence function from estimating $\pi$ with $\pi_n$. An analogous asymptotic expansion arises under alternative sampling schemes when $\nlab$ is fixed and hence the propensity score is effectively known. \cite{gan2023prediction, miao2023assumption} Lastly, of $\bphi(\Yhat,\bX,\bbeta^*)=\bomega^{\opt}(\Yhat,\bX)$, then $\E\left[\bphi(\Yhat,\bX;\bbeta^*)\right]=\bzero$ and the estimator would reach the semiparametric efficiency bound.}}

\section{The \texttt{CC} estimator}\label{more-eff-estimator}
\blue{Building on the findings of the previous section, we now propose an estimator whose influence function has an augmenting function that guarantees its asymptotic variance is never larger than that of classical inference. Unlike the extensions of \texttt{PPI}, which augment the empirical estimating equation for $\bbetahatlab$ with information from $\Yhat$, we propose to directly augment $\bbetahatlab$.\footnote{Angelopoulos et al.\cite{angelopoulos2023prediction} considered an augmented estimator for the special case of linear regression (see Algorithm S4 of the supplementary materials), which is generally distinct from the solution to (\ref{eq: ppi}).} Such augmented estimators have been widely studied in statistics and related areas as a means of improving efficiency, yet remain relatively underexplored in the context of PB inference (see Section~\ref{other-connection} for further discussion). In particular, within the two-phase sampling literature, Chen and Chen\cite{chen2000unified} (\texttt{CC}) proposed an estimator for the regression coefficients of a GLM that leverages a small validation dataset with gold-standard measurements together with a larger primary dataset containing less-expensive proxy measurements. The \texttt{CC} estimator adds an additional term to the classical estimator that can be estimated from the proxy measurements and that is consistent for zero. This term is carefully designed to be correlated with the classical estimator in a way that guarantees a variance reduction.}

\blue{To apply the this approach to the PB inference problem, let $\bW^\cc = \bC_{12}\bC_{22}^{-1}\bB^{-1}$ where $\bC_{12} = \text{Cov}[\bphi(Y, \bX; \bbeta^*), \bphi(\Yhat, \bX;\bgamma^*)]$, $\bC_{22} = \text{Var}[\bphi(\Yhat, \bX;\bgamma^*)]$,
and $\bB = -\E\left[ \frac{\partial \bphi(\Yhat, \bX; \bgamma^*)}{\partial \bgamma^{\top}} \right]^{-1}$. We then define the \texttt{CC} estimator as
$$\bbetahatcc = \bbetahatlab  - \bAhat \widehat{\bW}^{\cc}\left(\bgammahatlab  - \bgammahatall  \right).$$ 
where $\bgammahatlab$ is the solution to $n^{-1}\sum_{i=1}^n\bphi(\Yhat,\bX_i;\bgamma)\frac{R_i}{\pi_n}=\bzero$, $\widehat{\bA}$ is a consistent estimator of $\bA$, and $\widehat{\bW}^{\cc}$ is a consistent estimator of $\bW^\cc$. Proposition \ref{prop: betacc} shows that $\bbetahatcc$ is asymptotically linear with an influence function resembling that of $\bbetahatppi$, but with a different augmenting function. This result was previously shown for GLMs in the original paper by Chen \& Chen.\citep{chen2000unified} We present it here applied to the PB inference problem as a reference.}

\blue{\begin{prop}[Asymptotic linearity of $\bbetahatcc$]\label{prop: betacc}
Under the assumptions in the Appendix, $\bbetahatcc$ is asymptotically linear such that:
\begin{align*}
    n^{1/2}(\bbetahatcc -\bbeta^*) = n^{-1/2}\sum_{i=1}^n {\bf{IF}}^{\cc}(Y_i, \Yhat_i,  \bX_i; \bbeta^*,\bgamma^*) + o_p(1),
\end{align*}
where ${\bf{IF}}^{\cc}(Y_i, \Yhat_i,  \bX_i; \bbeta^*,\bgamma^*) = \bA\left\{\bphi(Y_i,\bX_i;\bbeta^*)\frac{R_i}{\pi} + \bW^{\cc} \bB \bphi(\Yhat_i,\bX_i;\bgamma^*)(1-\frac{R_i}{\pi})\right\}$.
\end{prop}
$\bbetahatcc$ is therefore an AIPW estimator with
\begin{align*}\label{omega-cc}
    \bomega^\cc(\Yhat, \bX)  =  \text{Cov}[\bphi(Y, \bX; \bbeta^*), \bphi(\Yhat, \bX;\bgamma^*)] \{\text{Var}[\bphi(\Yhat, \bX;\bgamma^*)]\}^{-1}\bphi(\Yhat, \bX; \bgamma^*).
\end{align*}
The augmenting function is the orthogonal projection of $\bphi(Y, \bX; \bbeta^*)$ onto the subspace spanned by $\bphi(\Yhat, \bX; \bgamma^*)$ in the Hilbert space of mean-zero, square-integrable random variables. This projection effectively approximates $\bomega^{\opt}(\Yhat, \bX)$ as a linear combination of the components of $\bphi(\Yhat, \bX; \bgamma^*)$, yielding a closer approximation than $\bphi(\Yhat, \bX; \bgamma^*)$ alone and protecting against potential efficiency loss. Corollary \ref{coro: betacc} demonstrates that, unlike \texttt{PPI}, $\bbetahatcc$ is never asymptotically outperformed by $\bbetahatlab$.} 

\blue{\begin{corollary}[Efficiency of $\bbetahatlab$ vs. $\bbetahatcc$]\label{coro: betacc}
The difference in the asymptotic variance of 
$\bbetahatlab$ and $\bbetahatcc$ is $\Delta_{\var}( \bbetahatlab, \bbetahatcc) =  \bA\bC_{12}\bC_{22}^{-1}\bC_{12}\trans\bA\trans (\pi^{-1}-1)$. Consequently, $\bbetahatcc$ is at least as efficient as $\bbetahatlab$ and is more efficient if and only if $\bC_{12} \neq \bzero$.
\end{corollary}}

\blue{To understand Corollary~\ref{coro: betacc}, we consider the simple setting of homoskedastic linear regression to illustrate how the \texttt{CC} estimator provides efficiency gains.}

\vskip12pt\noindent\textbf{Example 1} (Homoskedastic Linear Regression). 
{\it Consider the setting of linear regression and let $\epsilon_{Y} = Y - \bX\trans\bbeta^*$ and $\epsilon_{\Yhat} = \Yhat - \bX\trans\bgamma^*$. Under homoskedasticity, $\bC_{12} = \sigma_{Y, \Yhat} \bSigma_{\bX}$ and $\bC_{22} = \sigma_{\Yhat}^2 \bSigma_{\bX}$ where $\bSigma_{\bX} = \E(\bX \bX\trans)$, $\sigma_{Y, \Yhat} = \E(\epsilon_{Y} \epsilon_{\Yhat})$ and $\sigma_{\Yhat}^2 = \E(\epsilon_{\Yhat}^2)$. Applying Corollary \ref{coro: betacc}, 
\begin{align*}
\Delta_{\var}(\bbetahatlab,   \bbetahatcc )  &=  \left(\pi^{-1} - 1\right) \left( \frac{ \sigma_{Y, \Yhat}^2 } {\sigma_{\Yhat}^2}\right) \bSigma_{\bX}^{-1}=  \left(\pi^{-1} - 1\right) \sigma_Y^2  \rho_{Y, \Yhat}^2\bSigma_{\bX}^{-1}.
\end{align*}
where $\rho_{Y, \Yhat} = \sigma_{Y, \Yhat} / \sigma_{Y} \sigma_{\Yhat}$ is the partial correlation of the residuals for $Y$ and $\Yhat$ given $\bX$. $\bbetahatppi$ is thus more efficient than $\bbetahatlab$ when $\Yhat$ is a strong predictor for $Y$ after adjusting for $\bX$, though $\Yhat$ does not necessarily need to be derived from a ML model that estimates the true regression function. The improvement is stronger with increasing residual correlation of the residuals between $Y$ and $\Yhat$, greater variability in the residuals for $Y$, and a lower probability of labeling.}

\blue{In addition to being more efficient than classical inference, the \texttt{CC} estimator has the lowest asymptotic variance among all weighted augmented estimators. This optimality property is suggested by the orthogonal projection interpretation and indicates that $\bAhat\bWhat^{\cc}$ is a reasonable choice of weights for the augmented estimator. This result was mentioned in the original paper by Chen \& Chen\citep{chen2000unified} and we provide a proof in the Appendix.} 

\begin{prop}[$\bbetahatcc$ is the most efficient among augmented estimators] \label{prop: betaCC} Let $\bbetatilde^{\cc}(\bVhat)=\bbetahatlab - \bVhat(\bgammahatlab-\bgammahatall)$ be an alternative weighted augmented estimator with weights $\bVhat$ that are consistent for some $\bV^* \in \mathbb{R}^{q\times q}$. Then the difference in asymptotic variance is such that $\Delta_{\var}(\bbetatilde^{\cc}(\bVhat), \bbetahatcc) \succeq \bzero$.
\end{prop}

\subsection{Relationship to the \texttt{PPI} extensions}
\blue{We now return to the extensions of \texttt{PPI} introduced in Section \ref{motwani-recap}, which have a similar goal of improving the efficiency of classical inference. We focus on the \texttt{PDC} estimator as it is known to be the most efficient among estimators that are solutions to the weighted estimating equation in (\ref{eq: ppi-ext}).\cite{xu2025unified}. Specifically, let $\bW^\pdc = \bD_{12}\bD_{22}^{-1}$. $\bbetahatpdc$ is the solution to 
\begin{align*}
n^{-1} \sum_{i = 1}^{n} \bphi(Y_i, \bX_i; \bbeta) \frac{R_i}{\pi_n}  -  \bWhat^{\pdc} \left[n^{-1} \sum_{i = 1}^{n}  \bphi(\Yhat_i, \bX_i; \bbeta) \frac{R_i}{\pi_n} -  n^{-1} \sum_{i = 1}^{n}  \bphi(\Yhat_i, \bX_i; \bbeta) \right] = \bzero
\end{align*}
where $\bWhat^{\pdc}$ is a consistent estimator of $\bW^\pdc$. We show in Proposition \ref{prop: betapdc} that, similar to the \texttt{PPI} and \texttt{CC} estimators, $\bbetahatpdc$ is asymptotically linear with an AIPW influence function. The asymptotic normality of the \texttt{PDC} estimator has been shown in Gan et al.\cite{gan2023prediction} when the labeled and unlabeled data are of fixed size.}

\begin{prop}[Asymptotic linearity of $\bbetahatpdc$] \label{prop: betapdc}
Under the basic assumptions in the Appendix, $\bbetahatpdc$ is consistent such that $||\bbetahatpdc -\bbeta^*|| = o_p(1)$. Moreover, $\bbetahatpdc$ is asymptotically linear such that:
\begin{align*}
    n^{1/2}(\bbetahatpdc-\bbeta^*) = n^{-1/2}\sum_{i=1}^n {\bf{IF}}^{\pdc}(Y_i, \Yhat_i,  \bX_i; \bbeta^*) + o_p(1),
\end{align*}
where ${\bf{IF}}^{\pdc}(Y_i, \Yhat_i,  \bX_i; \bbeta^*) = \bA\left[\bphi(Y_i,\bX_i;\bbeta^*)\frac{R_i}{\pi} + \bW^{\pdc} \left\{\bphi(\Yhat_i,\bX_i;\bbeta^*)- \E \left[\bphi(\Yhat_i,\bX_i;\bbeta^*) \right]\right\}(1-\frac{R_i}{\pi})\right]$.
\end{prop}

\blue{Akin to the \texttt{CC} estimator, the \texttt{PDC} estimator approximates $\bomega^{\opt}(\Yhat, \bX)$ with an orthogonal projection of $\bphi(Y, \bX; \bbeta^*)$ onto the subpace spanned by $\bphi(\Yhat, \bX; \bbeta^*) - \E \left[\bphi(\Yhat_i,\bX_i;\bbeta^*) \right]$ to protect against potential efficiency loss. The key difference between \texttt{CC} and \texttt{PDC} is that \texttt{PDC} does not estimate a separate $\bgamma^*$ for the  estimating equation in the augmenting function. It is not clear from analytic comparisons whether \texttt{CC} or \texttt{PDC} is more efficient in general due to this difference. In Proposition \ref{prop: betacc-betapdc-glm}, we show that in the specific case of GLMs with correctly specified models for $Y$ and $\Yhat$ given $\bX$, \texttt{CC} is more efficient than \texttt{PDC}. We evaluate this result numerically in our simulation studies in Section \ref{empirical-studies}.}

\blue{\begin{prop}[Efficiency of $\bbetahatcc$ vs.\ $\bbetahatpdc$ for GLMs] \label{prop: betacc-betapdc-glm} 
Suppose that we seek to estimate regression parameters from a GLM with canonical link such that $\bphi(y,\bx;\bbeta) = \bx\left\{y-g(\bx\trans\bbeta)\right\}$ with $g(\cdot)$ being the inverse link function. A sufficient condition for $\bbetahatcc$ to be at least as efficient as $\bbetahatpdc$ such that $\Delta_{\var}( \bbetahatpdc, \bbetahatcc)\succeq \bzero$ is for the conditional means of both $Y$ and $\Yhat$ to be correctly specified such that:
\begin{align*}
    \E(Y \mid \bX) = g(\bX\trans\bbeta^*) \quad \mbox{and} \quad \E(\Yhat \mid \bX) = g(\bX\trans\bgamma^*).
\end{align*}
\end{prop}}

\section{Related literature}\label{other-connection}
Before evaluating the empirical properties of the \texttt{CC} estimator, we provide a brief overview of several closely related areas. Existing PB inference literature has noted connections with semi-supervised inference, missing data, and measurement error. \citep{angelopoulos2023prediction, angelopoulos2023ppi++, miao2023assumption, zrnic2024active, zrnic2024cross, angelopoulos_note_2024, ji2025predictions} \blue{We provide a brief review of three areas that directly relate to the PB inference problem.} These connections underscore that aspects of this modern problem have been previously examined, albeit with differences in motivation, terminology, and notation. Given the close relationship between the \texttt{PPI} and \texttt{CC} methods, ongoing research in PB inference would benefit from a more comprehensive review of these and other related areas.

\subsection{Seemingly unrelated regression}
\blue{The augmentation-based approach used by the \texttt{CC} estimator is related to several other estimators developed for linear regression, both historically and in recent work. In particular, it connects to seemingly unrelated regression (SUR), a framework introduced by economist Arnold Zellner in 1962 for the joint estimation of multiple related regression models \citep{zellner1962efficient, swamy1975bayesian, schmidt1977estimation}.} In the current setting, the SUR problem can be written as 
\begin{equation*}
Y = \bX\trans \bbeta_0 + \epsilon \quad  \mbox{and} \quad  \Yhat = \bX\trans \bgamma_0 + \epsilon' \quad 
\end{equation*}
where $\E(\epsilon \mid \bX) = 0$ and $\E(\epsilon' \mid \bX) = 0.$ Letting  $W^{\sur}=  \frac{\E[\text{Cov}( \epsilon, \epsilon' \mid \bX )]}{\E[\text{Var}(\epsilon' \mid \bX)]}$, Conniffe \cite{conniffe1985estimating} proposed to estimate $\bbeta_0$ with:  
$$\bbetahatsur= \bbetahatlab  - \widehat{W}^{\sur}\left(\bgammahatlab - \bgammahatall \right) \quad \mbox{where} \quad  \widehat W^{\sur}=  \frac{\sum_{i = 1}^{\nlab} (Y_i - \bX_i\trans \bbetahatlab) (\Yhat_i - \bX_i\trans \bgammahatlab) }{\sum_{i = 1}^{\nlab} (\Yhat_i - \bX_i\trans \bgammahatlab)^2}.$$
In the homoskedastic case where $\text{Var}(\epsilon \mid \bX)$ and $\text{Cov}(\epsilon, \epsilon' \mid \bX)$ are constant and do not depend on $\bX$, $\bbetahatcc$ is asymptotically equivalent to $\bbetahatsur$. \citep{chen2000unified} Moreover, when the conditional distribution of $(\epsilon, \epsilon')$ given $\bX$ is a \blue{zero-mean bivariate normal distribution with constant variance-covariance, $\bbetahatsur$, and thus $\bbetahatcc$ as well, is the maximum likelihood estimator of $\bbeta_0$ and fully efficient under the assumed parametric model.\citep{conniffe1985estimating} Moreover, the estimator still provides valid inference when the outcomes are Missing at Random (MAR).} In other scenarios, $\bbetahatcc$ achieves a lower asymptotic variance than $\bbetahatsur$ due to Proposition \ref{prop: betaCC}. 

The SUR estimator was recently applied in a PB inference problem inspired by genome-wide association studies conducted with population biobanks. These studies often rely on pre-trained models to impute outcomes of interest (i.e., phenotypes or traits) as they are partially missing due to the time and expense of ascertainment. \citep{mccaw2024synthetic}  The proposed method, \texttt{SynSurr}, assumes the true and predicted outcomes follow a bivariate normal distribution. The \texttt{SynSurr} estimator is equivalent to $\bbetahatsur$ and asymptotically equivalent to $\bbetahatcc$ when $\text{Var}(\epsilon\mid\bX)$ and $\text{Cov}(\epsilon, \epsilon'\mid\bX)$ are constant. Building on \texttt{SynSurr}, Miao et al \cite{miao2024popgwas} proposed the \texttt{POP-GWAS} estimator for a variety of regression models utilizing the augmentation technique. \blue{The \texttt{POP-GWAS} estimator\footnote{The original \texttt{POP-GWAS} estimator used $\bgammahatunlab$ for the augmentation term. We updated the weight and augmentation term for consistency with the \texttt{CC} estimator.} is 
$$
\bbetahatpop = \bbetahatlab -  \widehat{W}^{\pop} \left(\bgammahatlab - \bgammahatall\right) \quad \mbox{where} \quad  \widehat{W}^{\pop} = \widehat{\text{Corr}}(Y, \Yhat \mid \bX).
$$
The weight, $\widehat{W}^\pop$, is not generally a consistent estimator of  $W^{\text{SUR}}$. $\bbetahatpop$ is therefore less efficient than $\bbetahatsur$ when the residuals are bivariate normal with constant variance-covariance and less efficient than $\bbetahat^{\cc}$ in general.}

\subsection{Augmented estimation for quantifying a treatment effect}
{\blue{Outside of linear regression,}} the augmented estimation approach has also been widely used for treatment effect estimation.\cite{leon2003semiparametric, lu2008improving, tsiatis2008covariate, gilbert2009efficient, zhang2010increasing, tian2012covariate, yang2020combining}  Many of these approaches build on methods from  survey sampling.\cite{cassel1976some, sarndal2003model} For example, Tsiatis et al \cite{tsiatis2008covariate} consider covariate adjusted analyses for comparing two treatments in randomized clinical trials. \blue{The goal is to estimate the treatment effect $\Delta = \mathbb{E}(Y \mid X = 1) - \mathbb{E}(Y \mid X = 0)$.} A standard estimator of $\Delta$ is $\widehat{\Delta} = \overline{Y}^1 - \overline{Y}^0$, where $\overline{Y}^1 = n_1^{-1} \sum_{i = 1}^n Y_i X_i$ and $\overline{Y}^0 = n_0^{-1} \sum_{i = 1}^n Y_i (1-X_i)$ are the empirical means of $Y$ in the treated and control groups of sizes $n_1$ and $n_0$, respectively. To utilize information from baseline covariates $\bZ$ for improving efficiency, Tsiatis et al \cite{tsiatis2008covariate} propose an augmented estimator of the treatment effect given by
\begin{equation}\label{treatment-EE}
\widehat{\Delta} -  \sum_{i = 1}^n (X_i - \bar{X}) [n_0^{-1} h^{(0)}(\bZ_i) + n_1^{-1} h^{(1)}(\bZ_i)],
\end{equation}
where $\bar{X} = n^{-1} \sum_{i = 1}^n X_i$ and  $h^{(k)}(\bZ)$ are arbitrary scalar functions of $\bZ$ for $k = 0, 1$. As $X$ and $\bZ$ are independent due to randomization, the augmentation term in \ref{treatment-EE} converges in probability to 0, analogously to the augmentation term in the \texttt{CC} estimator. This technique has been applied in numerous settings, most recently for the analysis of time-to-event outcomes\citep{zhang2025unified} and recurrent events.\citep{gronsbell2024nonparametric}

\subsection{Control variates in Monte Carlo simulations}
Lastly, the augmentation technique has been utilized to improve the efficiency of Monte Carlo simulations. Rothery \cite{rothery1982} considered using a \textit{control variate}, a statistic whose properties are known, to improve precision when estimating power via simulation. In particular, suppose interest lies in estimating the power of a test $T$ of $H_{0}: \theta = 0$ for some scalar parameter $\theta$. Let $S$ denote an existing test of the same hypothesis whose power is known. Define $Y$ as an indicator that test $T$ rejects, $Z$ as an indicator that test $S$ rejects, and let $\mu_{Z} = \mathbb{E}_{\theta}(Z)$ denote the power of reference test $S$. For a set of $n$ simulation replicates, the baseline estimator of power for $T$ is simply $\overline{Y} = n^{-1}\sum_{i=1}^{n}Y_{i}$ while a control variate augmented estimator is given by:
\begin{align}\label{rotherty-contvar}
 \frac{1}{n} \sum_{i=1}^n \left[ Y_i - \lambda \{Z_i - \mu_{Z}\}\right].
 \end{align}
The variance of the augmented estimator is minimized at $\lambda^{*} = \frac{\text{Cov}(Y, Z)}{\text{Var}(Z)}$. \citep{rothery1982, davidson1992}  The augmentation term in (\ref{rotherty-contvar}) reduces the variance of the unadjusted power estimate $\overline{Y}$ in proportion to the correlation between $Y$ and $Z$. Building on work from the survey sampling literature, a similar idea was later applied for semi-supervised estimation of means \cite{zhang2019semi} and for causal effect estimation with multiple observational data sources\cite{yang2020combining}.  

\section{Empirical studies}\label{empirical-studies}
\subsection{Data generating process}
\blue{We now return to the example of homoskedastic linear regression to build further intuition for the \texttt{CC} estimator and to facilitate explicit comparisons with the \texttt{PPI} and \texttt{PDC} estimators. Additional simulation results for Poisson regression are provided in the Appendix. Here we generated the outcome as
$$Y = \bX_1\trans \bbeta_1  + \bX_2\trans \bbeta_2 + \epsilon$$ 
where $\epsilon \sim N(0, \sigma^2_{\epsilon})$, $\epsilon \perp (\bX_1\trans, \bX_2\trans)\trans$, $\bX_j \sim \text{MVN}\left(\bm{0}, \sigma_j^2 \bI_d\right)$ for $j = 1, 2$, and $\bI_d$ denotes the $d \times d$ identity matrix. Our focus is on estimating 
$\bbeta_1^* \coloneq \E \left( \bX_1 \bX_1 \trans \right)^{-1} \E \left( \bX_1 Y \right)$, which equals $\bbeta_1$ in the current setting. For simplicity, we generated the prediction of $Y$ as
$$\Yhat = \bX_1\trans \btheta_1  + \bX_2\trans \btheta_2 + \bX_3\trans \btheta_3$$
where $\bX_3 \sim \text{MVN}\left(\bm{0}, \sigma_3^2 \bI_d\right)$ and $\bX_3 \perp (\epsilon, \bX_1\trans, \bX_2\trans)\trans$. We varied the values of $\btheta_j$ to control the prediction quality. The settings considered are summarized in Table \ref{tab: lin-reg-study}.}

\blue{Briefly, we considered two primary settings. The first is an ``ideal'' setting that mimics when the prediction is derived from a linear model that was estimated on a large, independent dataset with the same distribution of $Y \mid (\bX_1, \bX_2)$ as the dataset used for conducting inference on $\bbeta_1^*$. We consider two ideal sub-settings reflecting when either both $\bX_1$ and $\bX_2$ were available for model training or only $\bX_2$ was available. The second is a ``distributional shift'' setting that mimics the practical scenario in which the prediction model was trained on a dataset with a different distribution than the dataset used for inference. We consider four types of distributional shift.}

\begin{table}[ht]
\centering
\begin{tabular}{lccc l}
\toprule
\textbf{Setting} & $\btheta_1$ & $\btheta_2$ & $\btheta_3$ & \textbf{Description} \\
\midrule
\multicolumn{5}{l}{\textbf{Ideal}} \\
1 & $\bbeta_1$ & $\bbeta_2$ & $\mathbf{0}$ & $\widehat Y = \E(Y \mid \bX_1, \bX_2)$ \\
 2 & $\mathbf{0}$ & $\bbeta_2$ & $\mathbf{0}$ & $\widehat Y = \E(Y \mid \bX_2)$ \\
\midrule
\multicolumn{5}{l}{\textbf{Distributional Shift}} \\
1 & $\mathbf{0}$ & $\bbeta_2$ & $\btheta_3$ & Shift in the distribution of $Y \mid (\bX_1, \bX_3)$ \\
2 & $\btheta_1$ & $\bbeta_2$ & $\mathbf{0}$ & Shift in the distribution of $Y \mid \bX_1$ \\
3 & $\bbeta_1$ & $\btheta_2$ & $\mathbf{0}$ & Shift in the distribution of $Y \mid  \bX_2$ \\
4 & $\btheta_1$ & $\btheta_2$ & $\mathbf{0}$ & Shift in the distribution of $Y \mid (\bX_1, \bX_2)$ \\
\bottomrule
\end{tabular}
\caption{\blue{Description of the ideal and distributional shift settings for the linear regression simulation study.}}\label{tab: lin-reg-study}
\label{tab:simulation-settings}
\end{table}

\blue{In all settings, we compare the classical OLS estimator against \texttt{CC}, \texttt{PPI}, and \texttt{PDC}. Following Ji et al.\ \cite{ji2025predictions}, we obtained analytic expressions for the asymptotic variance of each PB method to identify the primary parameter controlling the efficiency gain in each sub-setting (see Appendix). We assessed numerical performance by generating 1,000 datasets for various values of this parameter and reporting the average width and coverage of the 95\% confidence intervals for the first component of $\bbeta_1^*$ for each method.}

\blue{Throughout, we set $d = 5$, $n = 10,000$, $\pi = 0.1$, and generated $\bbeta_1$ and $\bbeta_2$ uniformly from the unit sphere $\mathbb{S}^{d-1}$ in each replicate. In the two ideal sub-settings, we let $\sigma^2_{\bX_1} + \sigma^2_{\bX_2} = 20$ and assessed performance across the variance ratio $\sigma^2_{\bX_2}/ \sigma^2_{\bX_1}  \in [0.5,  2]$, which quantifies the signal strength of $\bX_2$ for $Y$ after adjusting for $\bX_1$. In the distributional shift scenarios, we let $\sigma^2_{\bX_1} = \sigma^2_{\bX_2} = 10$. For sub-setting 1, we generated $\btheta_3$ uniformly from the unit sphere $\mathbb{S}^{d-1}$ in each replicate and assessed performance across $\sigma^2_{\bX_3} / \sigma^2_{\bX_2} \in [0.5, 2]$, which quantifies the degree of spurious signal in $\Yhat$ relative to the true signal from $\bX_2$. For subsettings 2-4, we generated $\dfrac{\bbeta_j - \btheta_j}{\| \bbeta_j - \btheta_j \|}$ uniformly from the unit sphere $\mathbb{S}^{d-1}$ in each replicate and assessed performance across $\| \bbeta_j - \btheta_j \| \in [0, 2]$, which measures the magnitude of distributional shift.}

\subsection{Results}
\subsubsection{Ideal scenario}
\begin{figure}[H]
    \centering
    \begin{subfigure}[a]{\textwidth}
        \centering
        \includegraphics[width=\textwidth]{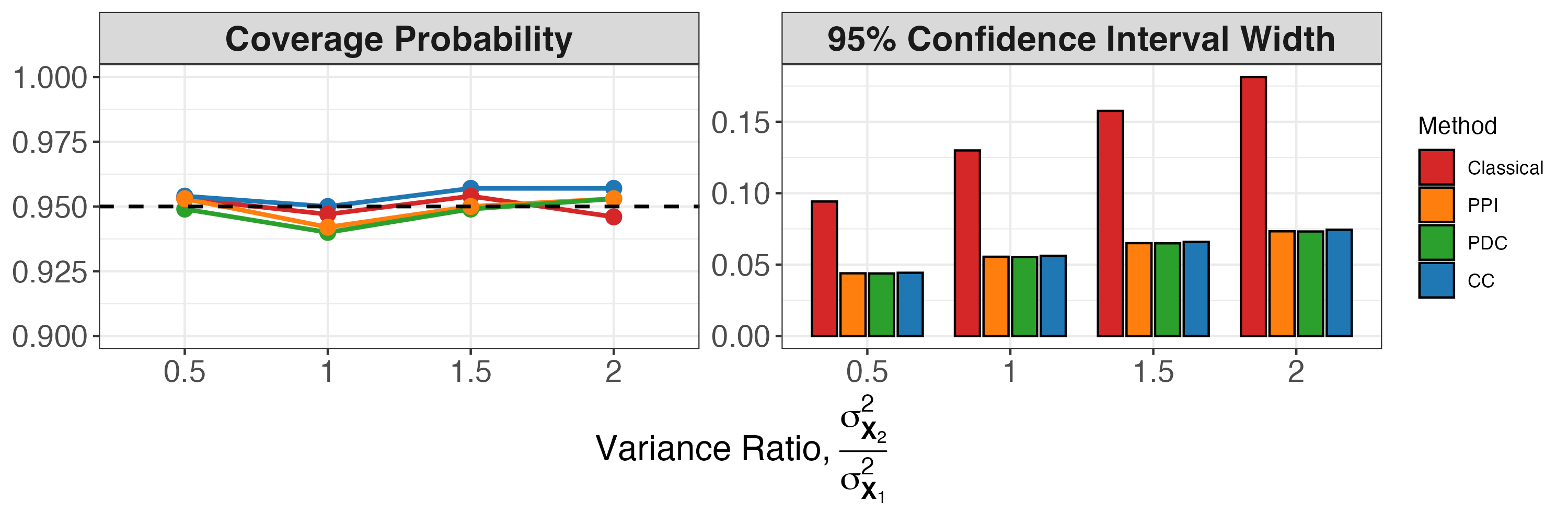}
        \caption{}
          \label{fig: fo-results}
    \end{subfigure}

    \begin{subfigure}[b]{\textwidth}
        \centering
        \includegraphics[width=\textwidth]{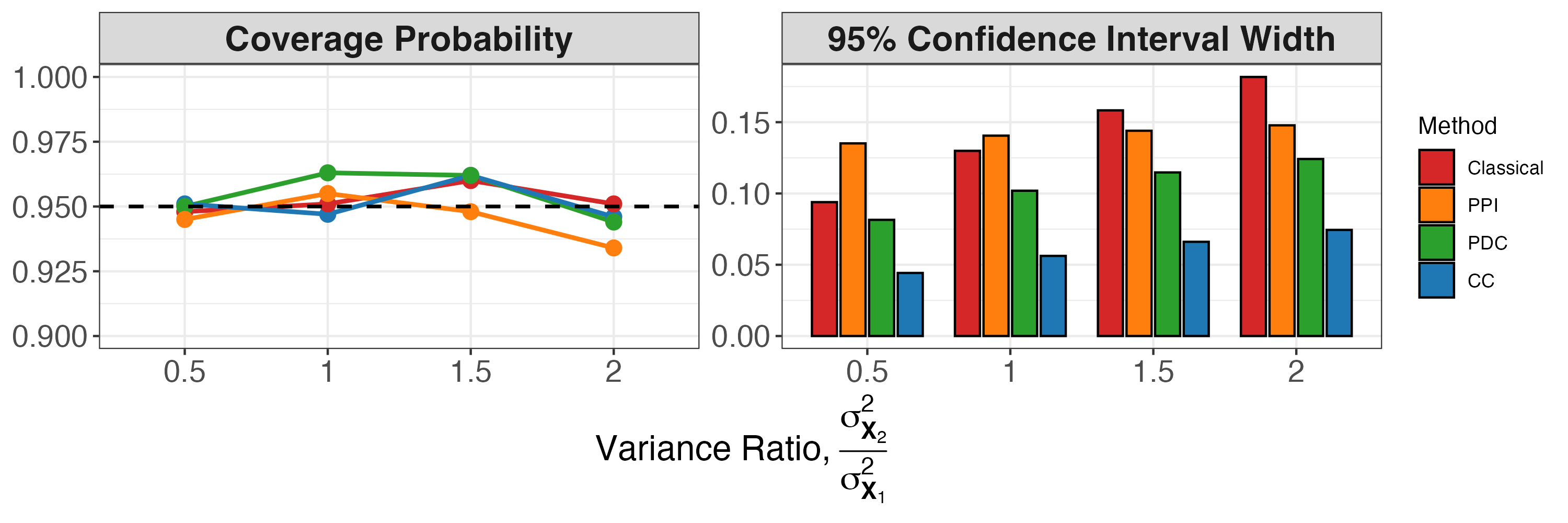}
        \caption{}
          \label{fig: po-results}
    \end{subfigure}

    \begin{subfigure}[c]{\textwidth}
        \centering
        \includegraphics[width=\textwidth]{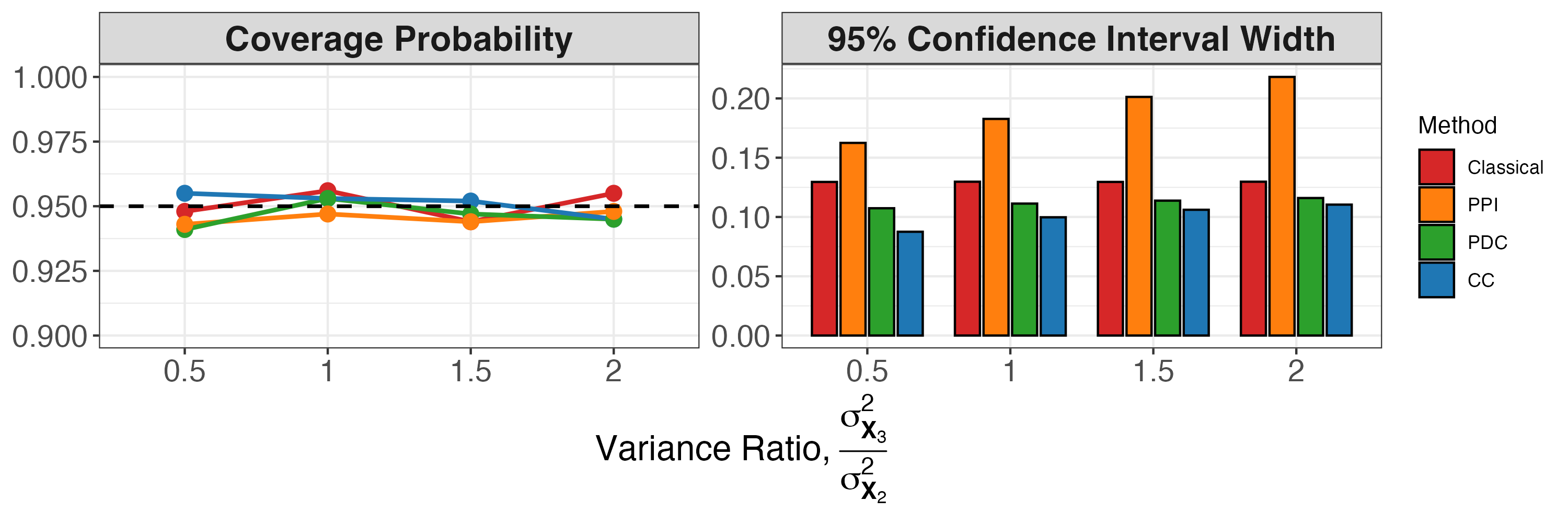}
        \caption{}
          \label{fig: np-results}
    \end{subfigure}

    \caption{Coverage probability and width of the 95\% confidence intervals for the ideal simulation scenario with (a) $\Yhat = \E(Y \mid \bX_1, \bX_2) = \bX_1\trans \bbeta_1 + \bX_2\trans \bbeta_2$ or (b) $\Yhat = \E(Y \mid \bX_2) = \bX_2\trans \bbeta_2$ and the distributional shift scenario with (c) $\Yhat = \bX_2\trans \bbeta_2 + \bX_3\trans \btheta_3$.}
    \label{fig: results-1}
\end{figure}

\blue{The results for the ideal scenario with $\Yhat = \E(Y \mid \bX_1, \bX_2)$ are presented in Figure \ref{fig: fo-results} while those with $\Yhat = \E(Y \mid \bX_2)$ are presented in Figure \ref{fig: po-results}. In the former case, all three approaches are asymptotically equivalent and significantly outperform the classical OLS estimator as they achieve the semi-parametric efficiency bound. The efficiency improvement increases as $\bX_2$ explains more of the variation in $Y$ than $\bX_1$ (i.e., as $\sigma^2_{\bX_2}/ \sigma^2_{\bX_1}$ increases). Intuitively, we expect \texttt{CC} and \texttt{PDC} to be equivalent as $\bgamma_1^* = \E\left( \bX_1 \bX_1 \trans  \right) \E\left[ \bX_1  \E(Y \mid \bX_1, \bX_2)  \right]  = \bbeta_1$. Moreover, the data-adaptive weighting employed by these methods becomes unnecessary when the optimal prediction is used, so \texttt{PPI} performs equally as well as \texttt{CC} and \texttt{PDC}. In the latter case with $\Yhat = \E(Y \mid  \bX_2)$, \texttt{PDC} and \texttt{PPI} lose efficiency as $\bgamma_1^* \ne \bbeta_1^*$. Relative to the first sub-setting, \texttt{CC} is unimpacted by the removal of $\bX_1$ from $\Yhat$ as its efficiency is determined by the residual correlation of $\Yhat$ for $Y$ after adjusting for $\bX_1$ (see Example 1). Additionally, this is a scenario in which \texttt{PPI} is outperformed by classical estimation, specifically when $\bX_1$ explains more of the variation in $Y$ than $\bX_2$ (i.e., when $\sigma^2_{\bX_2}/ \sigma^2_{\bX_1} < 1$). Lastly, the coverage of the intervals for all methods is close to 0.95 in both sub-settings.}  

\newpage

\subsubsection{Distributional Shift Scenario}
\blue{Starker differences in performance are observable in the distributional shift scenario. Figure \ref{fig: np-results} presents the results for the first sub-setting where $\Yhat = \E(Y \mid \bX_2) + \bX_3\trans \btheta_3$. \texttt{PPI} is substantially outperformed by classical estimation, particularly when the the variance of $\bX_3$, which is uninformative of $Y$, subsumes that of $\bX_2$ (i.e., $\sigma^2_{\bX_3}/\sigma^2_{\bX_2} > 1$). In contrast, the \texttt{CC} and \texttt{PDC} estimators still provide efficiency gains, with \texttt{CC} estimator substantially outperforming the \texttt{PDC} estimator with $\sigma^2_{\bX_3}/\sigma^2_{\bX_2} < 1$.} 

\begin{figure}[H]
    \centering
    \begin{subfigure}[b]{\textwidth}
        \centering
        \includegraphics[width=\textwidth]{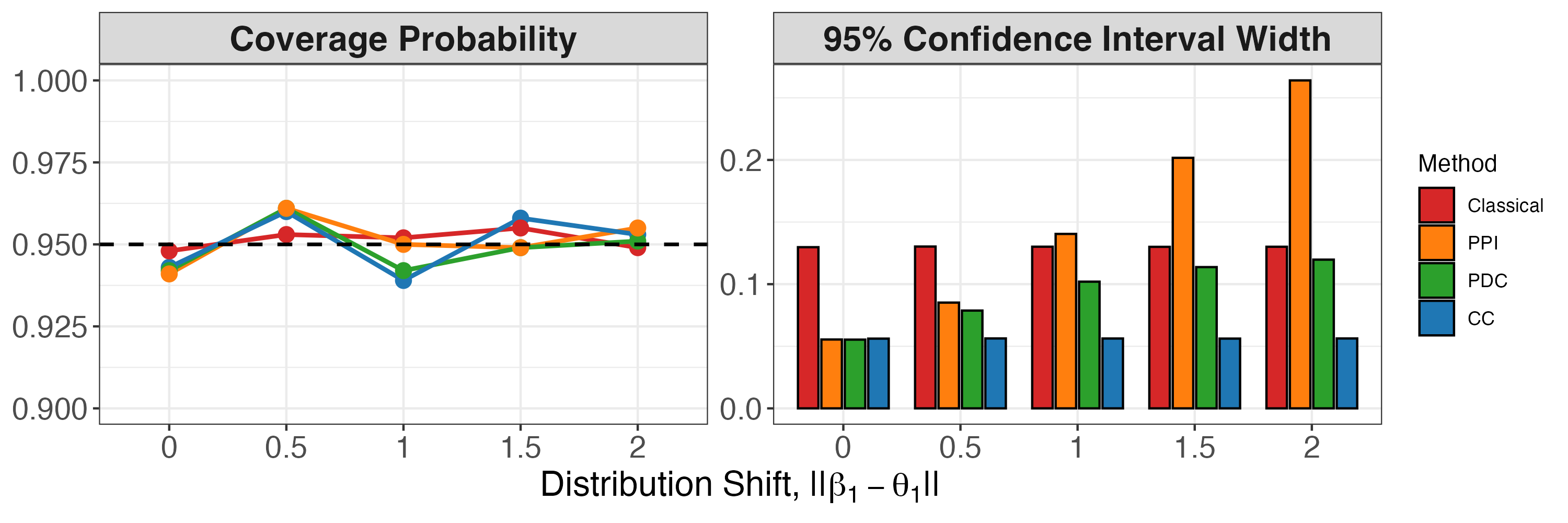}
        \caption{}
          \label{fig: x1s-results}
    \end{subfigure}

    \begin{subfigure}[b]{\textwidth}
        \centering
        \includegraphics[width=\textwidth]{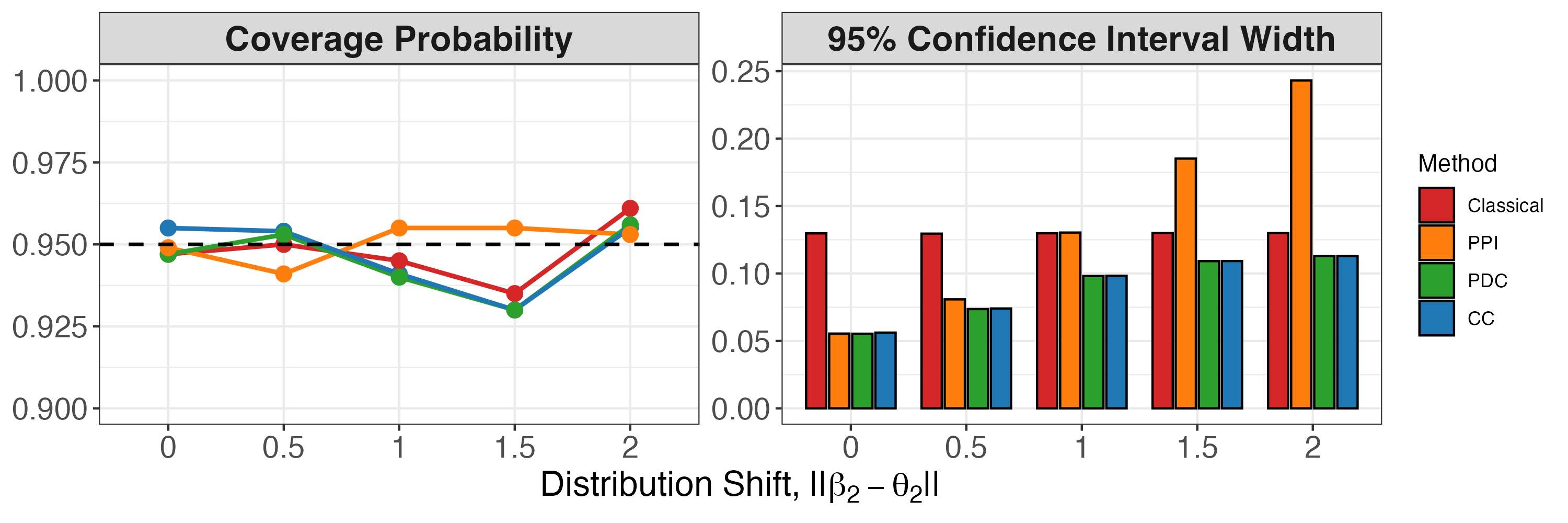}
        \caption{}
          \label{fig: x2s-results}
    \end{subfigure}

     \begin{subfigure}[a]{\textwidth}
        \centering
        \includegraphics[width=\textwidth]{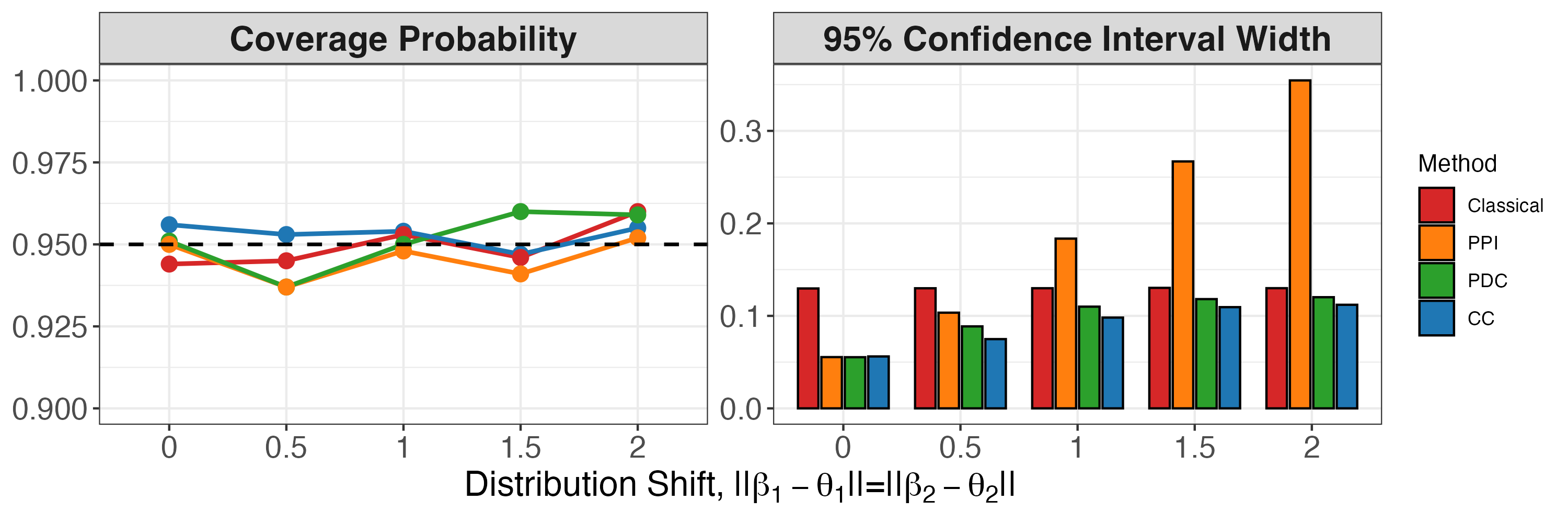}
        \caption{}
          \label{fig: x1x2s-results}
    \end{subfigure}
    
    \caption{Coverage probability and width of the 95\% confidence intervals for the distributional shift simulation scenario with (a) $\Yhat = \bX_1\trans \btheta_1 + \bX_2\trans \bbeta_2$, (b) $\Yhat = \bX_1\trans \bbeta_1 + \bX_2\trans \btheta_2$ or (c) $\Yhat = \bX_1\trans \btheta_1 + \bX_2\trans \btheta_2$}
    \label{fig: dist-shift-results-1}
\end{figure}

\blue{The results for the remaining three distributional shift settings are presented in Figure \ref{fig: x1s-results} - \ref{fig: x1x2s-results}. Generally, \texttt{PPI} is outperformed by classical OLS as the magnitude of shift increases. For the sub-setting with $\Yhat = \bX_1\trans \btheta_1 + \bX_2\trans \bbeta_2$ presented in Figure \ref{fig: x1s-results}, the \texttt{CC} estimator is unimpacted by the magnitude of $\| \bbeta_1 - \btheta_1 \|$. In contrast, \texttt{PDC} performs similarly to classical OLS as $\| \bbeta_1 - \btheta_1 \|$ increases. For the sub-setting with $\Yhat = \bX_1\trans \bbeta_1 + \bX_2\trans \btheta_2$, however, \texttt{CC} and \texttt{PDC} are again equivalent as $\bgamma_1^* = \bbeta_1^*$. Lastly, in the sub-setting with $\Yhat = \bX_1\trans \btheta_1 + \bX_2\trans \btheta_2$ in Figure \ref{fig: x1x2s-results}, the \texttt{CC} and \texttt{PDC} methods both offer substantial improvements over OLS with moderate shift, with \texttt{CC} again outperforming \texttt{PDC}. Similar to the ideal setting, all methods achieve coverage close to the nominal level across all sub-settings. Overall, the \texttt{CC} estimator provides the highest efficiency gains across all settings considered, which underscores the result in Proposition \ref {prop: betacc-betapdc-glm}.}

\newpage

\section{UK Biobank Analysis}\label{real-data}
We also evaluate the performance of the PB inference methods in an analysis focused on central obesity. Central obesity is excessive accumulation of body fat around the abdominal region and is strongly associated with cardiometabolic diseases such as type 2 diabetes and cardiovascular disease. A major challenge in its study is the limited availability of direct body composition measurements. Android fat mass (AFM), measured via a dual-energy X-ray absorptiometry (DEXA) scan, is a key marker for assessing central obesity in clinical practice. \citep{zhu_dxa-derived_2022} However, due to its high cost, DEXA-derived AFM is usually only measured in a small subset of participants in large-scale studies. For example, in the UK Biobank,\cite{sudlow_uk_2015} a prospective cohort study measuring genetic, lifestyle, and health information on approximately 500,000 participants in the UK, DEXA-derived AFM is incompletely ascertained, particularly among populations historically underrepresented in genetic research. \citep{sudlow_uk_2015} For instance, out of $5,397$ South Asian participants in the UK Biobank, only $288$ (5.3\%) participants have undergone DEXA scans. 

Our goal is to examine \blue{the association of AFM with age, sex, and body mass index} (BMI) among the South Asian population. Previous research has indicated that this population is at higher risk of central obesity and that it is associated with cardiometabolic diseases at lower BMI thresholds than European populations. \citep{sniderman_why_2007} Since DEXA-derived AFM is missing for most South Asian participants, it provides a natural opportunity to leverage ML-predicted AFM in downstream analyses. To illustrate the impact of different ML models on the efficiency of the PB inference methods, we trained two random forest models in the White British population ($32,460$ participants) and used them to predict missing AFM values in the South Asian population. We built a ``strong'' prediction model that incorporated both the covariates of interest (i.e., age, sex, and BMI) as well as additional predictors, including smoking status, alcohol consumption, two bioelectrical impedance measurements of the legs, physical activity scores from the International Physical Activity Questionnaire, and sedentary behavior quantified as total hours spent driving, using a computer, and watching television. We built another ``weak'' prediction model that included only the additional predictors. The strong prediction model achieved an $R^2$ of 0.8 while the weak prediction model had a substantially lower $R^2$ of 0.5 in the South Asian subpopulation.

Table \ref{tab:real-data} presents point estimates, standard errors, and the confidence interval ratio (CIR), defined as the \blue{ratio of the standard error of the given estimator to that of classical OLS, for the two prediction model settings.} Echoing our theoretical and simulation study results, all methods yield similar point estimates as they are all consistent estimators of the same parameter. In the analysis with the strong prediction model, all three PB inference outperform the classical estimator. \blue{Relative to \texttt{PPI} and \texttt{PDC}, \texttt{CC} provides an additional 11\% reduction in standard error for the coefficient for sex and a 20\% reduction for the coefficient for BMI. In the analysis with the weak prediction model, \texttt{PPI} yields larger standard errors than classical estimation for all parameters of interest, with a CIR of at least 1.35 for all of the coefficients. \blue{In contrast, \texttt{CC} and \texttt{PDC} are not outperformed by classical estimator and even offer standard error reductions of 22\% and 13\%, respectively, for the coefficient for BMI. $\bbetahatcc$ also provides a small 7\% reduction for the coefficient for sex.} Overall, the \texttt{CC} estimator is the best performing method in terms of efficiency in this example.}

\section{Discussion}\label{discussion}
\blue{We introduced the \texttt{CC} approach for Z-estimation problems with ML-imputed outcomes, which yields valid inference regardless of prediction quality while also achieving the lowest variance among weighted augmented estimators. Our theoretical and empirical results demonstrate that the \texttt{CC} estimator can provide practical improvements over classical estimation as well as the \texttt{PPI} estimator and its extensions, making it a preferred approach for applied PB inference tasks. More broadly, our work situates PB inference within semi-parametric efficiency theory and highlights connections to classical methods in statistics and related fields that are directly relevant to this modern problem.}

\blue{For simplicity and to facilitate comparisons with existing PB inference methods, we assumed that the outcomes are MCAR. Extending the \texttt{CC} approach to accommodate more general missingness patterns would greatly enhance its practical applicability and has been a focus of recent work.\cite{chen2025unified, chen2025surrogate} Additionally, the \texttt{CC} estimator is not guaranteed to outperform \texttt{PDC} in general. For instance, the result in Proposition \ref{prop: betacc-betapdc-glm} does not hold when the conditional mean of $Y$ given $\bX$ follows a logistic regression model and $\Yhat$ are non-differentially misclassified versions of $Y$. In this case, the conditional mean of $\Yhat$ given $\bX$ involves a different link function that is dependent on its sensitivity and specificity for $Y$.\cite{neuhaus1999bias} However, both estimators are special cases of recently proposed semi-supervised estimators that approximate $\omega^{\opt}(\Yhat, \bX)$ with an orthogonal projection of $\bphi(Y,  X; \bbeta^*)$ onto a carefully constructed basis\cite{xu2025unified}, suggesting a natural pathway for further improvement.}

It is also important to note that the \texttt{CC} estimator requires estimation of a $q \times q$ weight matrix, which can be unstable when labeled data are limited. In such cases, restricting the weight matrix to a simple structure, such as a scalar or diagonal matrix akin to \texttt{PPI++} or \texttt{PSPA}, may be beneficial in finite samples, albeit at some cost of asymptotic efficiency. Finally, although we focused on ML-imputed outcomes, the \texttt{CC} framework can accommodate ML-imputed covariates \cite{kluger2025prediction} and can be extended to handle multiple predictions of $Y$.\cite{lu2024leveraging, shan2025sada}

\newpage
\bigskip
\begin{center}
{\large\bf SUPPLEMENTARY MATERIAL}
\end{center}
The supplementary materials include proofs for theorems.
\begin{description}
\item[Code availability:] \blue{Our analyses can be replicated with the code at: \url{https://github.com/jlgrons/ML-Assisted-Inference}.}

\item[UK Biobank Dataset:] This work used data from the UK Biobank study (\url{https://www.ukbiobank.ac.uk})
and our access was approved under application 64875.
\end{description}

\bigskip
\begin{center}
{\large\bf ACKNOWLEDGMENTS}
\end{center}
J. Gronsbell gratefully acknowledges funding from an NSERC Discovery Grant (RGPIN-2021-03734), the Connaught Fund, the University of Toronto Data Science Institute, and the McLaughlin Center for Genetic Research.

\vskip 0.2in

\bibliographystyle{amawiley}
\bibliography{sampleama}



\begin{table}[H]
\centering
\footnotesize
\begin{minipage}[t]{0.48\textwidth}
\begin{tabular}{|l|cccc|}
\hline
\multicolumn{5}{|c|}{\textbf{Strong Prediction Model}} \\ 
\hline
 & Intercept & Age & Sex & BMI \\
\hline
\textbf{$\bbetahatlab$} &&&& \\
EST  & -4.857 & -0.003 & 0.813 & 0.173 \\
SE   & 0.3080  & 0.0041  & 0.0702  & 0.0087 \\
CIR & 1.00  & 1.00  & 1.00  & 1.00 \\
\hline
\textbf{$\bbetahatppi$} &&&& \\
EST  & -4.528 & -0.007 & 0.901 & 0.165 \\
SE   & 0.2602  & 0.0036  & 0.0628  & 0.0070 \\
CIR & 0.85  & 0.85  & 0.89  & 0.81 \\
\hline
\textbf{$\bbetahatcc$} &&&& \\
EST  & -4.553  & -0.006 & 0.903 & 0.163 \\
SE   & 0.2590  & 0.0034 & 0.0570 & 0.0057 \\
CIR & 0.84  & 0.81  & {\bf{0.81}}  & {\bf{0.66}} \\
\hline
\textbf{$\bbetahatpdc$} &&&& \\
EST  & -4.241  & -0.009 & 0.946 & 0.155 \\
SE   & 0.2660  & 0.0036 & 0.0646 & 0.0072 \\
CIR & 0.86  & 0.86  & 0.92  & 0.83 \\
\hline
\end{tabular}
\end{minipage}
\begin{minipage}[t]{0.48\textwidth}
\begin{tabular}{|l|cccc|}
\hline
\multicolumn{5}{|c|}{\textbf{Weak Prediction Model}} \\ 
\hline
 & Intercept & Age & Sex & BMI \\
\hline
\textbf{$\bbetahatlab$} &&&& \\
EST  & -4.857 & -0.003 & 0.813 & 0.173 \\
SE   & 0.3080  & 0.0041  & 0.0702  & 0.0087 \\
CIR & 1.00  & 1.00  & 1.00  & 1.00 \\
\hline
\textbf{$\bbetahatppi$} &&&& \\
EST  & -4.557 & -0.006 & 0.836 & 0.166 \\
SE   & 0.4175  & 0.0060  & 0.0982  & 0.0118 \\
CIR & 1.35 & 1.45 & 1.41 & 1.35 \\
\hline
\textbf{$\bbetahatcc$} &&&& \\
EST  & -4.790 & -0.004 & 0.822 & 0.172 \\
SE   & 0.2941  & 0.0038  & 0.0652  & 0.0068 \\
CIR & 0.95 & 0.91 & {\bf{0.93}} & {\bf{0.78}} \\
\hline
\textbf{$\bbetahatpdc$} &&&& \\
EST  & -4.564 & -0.004 & 0.876 & 0.161 \\
SE   & 0.289 & 0.004 & 0.0723 & 0.0075 \\
CIR & 0.93 & 0.96 & 1.03 & 0.87 \\
\hline
\end{tabular}
\end{minipage}
\caption{\small \blue{UK Biobank analysis of android fat mass based on the classical OLS estimator ($\bbetahatlab$), the \texttt{PPI} estimator ($\bbetahatppi$), the \texttt{CC} estimator ($\bbetahatcc$), and the PDC estimator ($\bbetahatpdc$). Estimates (EST), standard errors (SE), and confidence interval ratio (CIR) are reported to evaluate performance.}}
\label{tab:real-data}
\end{table}

\end{document}


\def\spacingset#1{\renewcommand{\baselinestretch}%
{#1}\small\normalsize} \spacingset{1}


\if1\blind
{
  \title{\bf Appendix for \\
  Another look at statistical inference with \\ machine learning-imputed data}
  \author{Jessica Gronsbell$^{1}$\thanks{The authors gratefully acknowledge funding from an NSERC Discovery Grant (RGPIN-2021-03734), the Connaught Fund, the University of Toronto Data Science Institute, and the McLaughlin Center for Genetic Research.}\\
    and \\
    Jianhui Gao$^{1}$ \\
    and \\
    Zachary R. McCaw$^{2}$\\
    and\\
    Yaqi Shi$^{1}$\\
    and \\
    David Cheng$^{3}$\\
    $^{1}$Department of Statistical Sciences, University of Toronto, Toronto, ON\\
$^{2}$Department of Biostatistics, UNC Chapel Hill, Chapel Hill, NC\\
$^{3}$Biostatistics Center, Massachusetts General Hospital, Boston, MA}
  \maketitle
} \fi

\if0\blind
{
  \bigskip
  \bigskip
  \bigskip
  \begin{center}
    {\LARGE\bf Appendix for \\
  Another look at statistical inference with \\ machine learning-imputed data}
\end{center}
  \medskip
} \fi

\newpage
\spacingset{1.9} 

\section{Theoretical results}

\blue{For a matrix $\bM$, we take ${\bf{M}} \succeq {\bf{0}}$ and ${\bf{M}} \succ {\bf{0}}$ to mean that $\bM$ is positive semi-definite and positive definite, respectively. For a vector $\bv=(v_1,\ldots,v_p)\trans$,$||\bv ||=\left\{\sum_{j=1}^p v_{j}^2\right\}^{1/2}$ is the $\ell_2$-norm. For a matrix $\bM$, $||\bM||=\sigma_{max}(\bM)$ be the spectral norm, where $\sigma_{max}(\bM)$ denotes the largest singular value of $\bM$. For estimators $\bbetahat_1$ and $\bbetahat_2$ such that $n^{1/2}(\bbetahat_1-\bbeta^*)\overset{d}{\to}N(\bzero,\bSigma_1)$ and $n^{1/2}(\bbetahat_2-\bbeta^*)\overset{d}{\to}N(\bzero,\bSigma_2)$, we take $\Delta_{\var}(\bbetahat_1, \bbetahat_2)$ to be the difference in the asymptotic variance-covariance matrices such that $\Delta_{\var}(\bbetahat_1, \bbetahat_2) = \bSigma_1 - \bSigma_2$.
In the following, we detail basic assumptions and sketch the proofs for the theoretical results.}

\subsection{Basic assumptions}\label{subse:assump}
\noindent
\blue{The following assumptions are assumed throughout the theoretical derivations. The first condition ensures the existence of second moments of the estimating equations evaluated at $\bbeta^*$ and $\gamma^*$ to allow for basic results such as the central limit theorem. Assumption C.2-C.3 require parameters to be bounded and to be identified and are standard assumptions in Z-estimation. Assumptions C.4-C.6 are used to show uniform convergence of estimating equations and their derivatives. C.7 is used to ensure existence of inverse of estimating equations in expansions for showing asymptotic normality.}

\noindent
\blue{\begin{enumerate}[label=C.\arabic*]
    \item $\E||\bphi(Y,\bX;\bbeta^*)||^2<\infty$, $\E||\bphi(\Yhat,\bX;\bbeta^*)||^2<\infty$, and $\E||\bphi(\Yhat,\bX;\bgamma^*)||^2<\infty$.
    \item The population estimating equation solutions $\bbeta^*$ and $\bgamma^*$ are in compact sets in $\R^{q}$.
    \item $\E\bphi(Y,\bX,\bbeta)$ has a unique zero $\bbeta^*$ such that $\E\left\{ \bphi(Y,\bX,\bbeta)\right\} \neq \bzero$ if $\bbeta \neq \bbeta^*$. Similarly, $\E\bphi(\Yhat,\bX,\bgamma)$ has a unique zero $\bgamma^*$.
    \item $\bphi(Y,\bX;\bbeta)$ and $\bphi(\Yhat,\bX,\bgamma)$ are respectively continuously differentiable in $\bbeta$ and $\bgamma$ almost surely.
    \item $\bphi(Y,\bX,\bbeta)$ and $\bphi(\Yhat,\bX,\bgamma)$ are dominated by some integrable function $d_0(Y,\Yhat,\bX)$ such that $||\bphi(Y,\bX,\bbeta)||\leq d_0(Y,\Yhat,\bX)$ and $||\bphi(\Yhat,\bX,\bgamma)||\leq d_0(Y,\Yhat,\bX)$ for all $\bbeta,\bgamma$ and $\E\left\{ d_0(Y,\Yhat,\bX) \right\}<\infty$.
    \item $\frac{\partial}{\partial\bbeta}\bphi(Y,\bX,\bbeta)$ and $\frac{\partial}{\partial\bgamma}\bphi(\Yhat,\bX,\gamma)$ are dominated by some integrable function $d_1(Y,\Yhat,\bX)$ such that $||\frac{\partial}{\partial\bbeta}\bphi(Y,\bX,\bbeta)|| \leq d_1(Y,\Yhat,\bX)$ and $||\frac{\partial}{\partial\bgamma}\bphi(\Yhat,\bX,\bgamma)|| \leq d_1(Y,\Yhat,\bX)$ for all $\bbeta,\bgamma$ and $\E\left\{ d_1(Y,\Yhat,\bX)\right\} < \infty$.
    \item $\E\frac{\partial}{\partial\bbeta}\bphi(Y,\bX;\bbeta^*)$ and $\E \frac{\partial}{\partial\bgamma} \bphi(\Yhat,\bX;\bgamma^*)$ are non-singular.
\end{enumerate}}

\subsection{Results for \texttt{PPI}}
\blue{\begin{prop}[Asymptotic linearity of $\bbetahatppi$]\label{prop: betappi}
Under basic assumptions in {Appendix \ref{subse:assump}},  $\bbetahatppi$ is consistent such that $||\bbetahatppi-\bbeta^*|| = o_p(1)$. Moreover, it is asymptotically linear such that:
$$ n^{1/2} \left(\bbetahatppi - \bbeta^*\right) =  n^{-1/2}  \sum_{i=1}^n  {\bf{IF}}^{\ppi}(Y_i, \Yhat_i,  \bX_i; \bbeta^*)   + o_p(1),$$ 
where ${\bf{IF}}^{\ppi}(Y_i, \Yhat_i, \bX_i; \bbeta^*) =   \bA\left[\bphi(Y_i, \bX_i; \bbeta^*) \frac{R_i}{\pi} +  \left\{\bphi(\Yhat_i, \bX_i; \bbeta^*)-\E\bphi(\Yhat_i,\bX_i;\bbeta^*)\right\}(1-\frac{R_i}{\pi})  \right]$ and $\bA =  -\E\left\{ \frac{\partial}{\partial \bbeta\trans} \bphi(Y, \bX; \bbeta^*)\right\}^{-1}$.
\end{prop}}

\begin{proof}
\blue{Let $\bPsi_{n}(\bbeta) = n^{-1} \sum_{i=1}^n \bphi(Y_i,\bX_i;\bbeta^*)\frac{R_i}{\pi_n} + \bphi(\Yhat_i,\bX_i;\bbeta^*)(1-\frac{R_i}{\pi_n})$ be the estimating equation for the \texttt{PPI} estimator, and $\bPsi(\bbeta) = \E \bphi(Y_i,\bX_i;\bbeta)$. $\bPsi(\bbeta)$ has a unique zero by assumption. Moreover, we have:
\begin{align*}
\sup_{\bbeta}||\bPsi_n(\bbeta) - \bPsi(\bbeta)|| &\leq \sup_{\bbeta}||n^{-1}\sum_{i=1}^n \bphi(Y_i,\bX_i;\bbeta)\frac{R_i}{\pi} - \bPsi(\bbeta)|| \\
&\quad + \sup_{\bbeta}||n^{-1}\sum_{i=1}^n \bphi(Y_i,\bX_i;\bbeta) R_i - \E\{\bphi(Y_i,\bX_i;\bbeta) R_i\}|| \cdot |\pi_n^{-1} - \pi^{-1}| \\
&\quad + \sup_{\bbeta}||\E\{\phi(Y_i,\bX_i;\bbeta)R_i\}|| \cdot |\pi_n^{-1} - \pi^{-1}| \\
&\quad + \sup_{\bbeta}|| n^{-1}\sum_{i=1}^n \bphi(\Yhat_i,\bX_i;\bbeta) (1-\frac{R_i}{\pi})|| \\
&\quad + \sup_{\bbeta}||n^{-1}\sum_{i=1}^n\bphi(\Yhat_i,\bX_i;\bbeta)R_i-\E\{ \bphi(\Yhat_i,\bX_i;\bbeta)R_i\}|| \cdot |\pi^{-1}-\pi_n^{-1}|\\
&\quad + \sup_{\bbeta}||\E\{\bphi(\Yhat_i,\bX_i;\bbeta)R_i\}||\cdot |\pi^{-1}-\pi_n^{-1}|.
\end{align*}
Now since $\E\{\bphi(Y_i,\bX_i;\bbeta)\frac{R_i}{\pi}\} = \bPsi(\bbeta)$, the parameter space for $\bbeta$ is compact,  $\bphi(Y_i,\bX_i;\bbeta)$ is continuous in $\bbeta$, and $\bphi(Y_i,\bX_i;\bbeta)$ can be dominated by an integrable function, we have that $\sup_{\bbeta}||n^{-1}\sum_{i=1}^n \bphi(Y_i,\bX_i;\bbeta)\frac{R_i}{\pi} - \bPsi(\bbeta)|| = o_p(1)$ by Lemma 2.4 in \cite{newey1994large}. By similar arguments, we have that $\sup_{\bbeta}||n^{-1}\sum_{i=1}^n \bphi(Y_i,\bX_i;\bbeta) R_i - \E\{\bphi(Y_i,\bX_i;\bbeta) R_i\}||=o_p(1)$, $\sup_{\bbeta}||n^{-1}\sum_{i=1}^n\bphi(\Yhat_i,\bX_i;\bbeta)(1-\frac{R_i}{\pi})|| = o_p(1)$, and $\sup_{\bbeta}||n^{-1}\sum_{i=1}^n \bphi(\Yhat_i,\bX_i;\bbeta) R_i - \E\{\bphi(\Yhat_i,\bX_i;\bbeta) R_i\}|| = o_p(1)$. Additionally, $ \sup_{\bbeta}||\E\bphi(Y_i,\bX_i;\bbeta)R_i||$ and $ \sup_{\bbeta}||\E\bphi(\Yhat_i,\bX_i;\bbeta)R_i||$ are bounded since $\bphi(Y_i,\bX_i;\bbeta)$ and $\bphi(\Yhat_i,\bX_i;\bbeta)$ are dominated by integrable functions. As $|\pi_n^{-1}-\pi^{-1}| = o_p(1)$, by Slutsky's theorem, $\sup_{\bbeta}||\bPsi_n(\bbeta) - \bPsi(\bbeta)|| = o_p(1)$. Standard arguments (e.g., Theorem 2.1 in \cite{newey1994large}) then verify that $||\bbetahatppi-\bbeta^*||=o_p(1)$.}

\blue{By the mean value theorem, expanding $\bzero = \bPsi_n(\bbetahat)$ around $\bbeta^*$ yields $n^{1/2}(\bbetahatppi - \bbeta^*) = \bAhat n^{1/2}\bPsi_n(\bbeta^*)$, where $\bAhat = \left\{ -\frac{\partial}{\partial\bbeta\trans} \bPsi_n(\bbetatilde)\right\}^{-1}$ and $\bbetatilde$ is such that $||\bbetatilde-\bbeta^*||\leq ||\bbetahatppi -\bbeta^*||$. Applying similar arguments as those above, it can be shown that $\sup_{\bbeta}||\frac{\partial}{\partial\bbeta}\bPsi_n(\bbeta) - \E\frac{\partial}{\partial\bbeta}\bphi(Y_i,\bX_i;\bbeta)|| = o_p(1)$, using that the parameter space for $\bbeta$ is compact, $\frac{\partial}{\partial\bbeta}\bphi(Y_i,\bX_i;\bbeta)$ and $\frac{\partial}{\partial\bbeta}\bphi(\Yhat_i,\bX_i;\bbeta)$ are continuous in $\bbeta$, and $\frac{\partial}{\partial\bbeta}\bphi(Y_i,\bX_i;\bbeta)$ and $\frac{\partial}{\partial\bbeta}\bphi(\Yhat_i,\bX_i;\bbeta)$ are dominated by an integrable function. As $\frac{\partial}{\partial\bbeta}\bphi(Y_i,\bX_i;\bbeta)$ is continuous in $\bbeta$ and is dominated by an integrable function, by the dominated convergence theorem, $\E\frac{\partial}{\partial\bbeta}\bphi(Y_i,\bX_i;\bbeta)$ is continuous in $\bbeta$. Furthermore, as $||\bbetatilde-\bbeta^*||\leq ||\bbetahatppi -\bbeta^*||=o_p(1)$,  $||\bbetatilde-\bbeta^*|| = o_p(1)$, and, by continuous mapping theorem $||\E\frac{\partial}{\partial\bbeta}\bphi(Y_i,\bX_i;\bbetatilde) - \E\frac{\partial}{\partial\bbeta}\bphi(Y_i,\bX_i;\bbeta^*)|| = o_p(1)$. Thus $||\frac{\partial}{\partial\bbeta}\bPsi_n(\bbetatilde) - \E\frac{\partial}{\partial\bbeta}\bphi(Y_i,\bX_i;\bbeta^*)|| \leq \sup_{\bbeta}||\frac{\partial}{\partial\bbeta}\bPsi_n(\bbeta) - \E\frac{\partial}{\partial\bbeta}\bphi(Y_i,\bX_i;\bbeta)|| + ||\E\frac{\partial}{\partial\bbeta}\bphi(Y_i,\bX_i;\bbetatilde) - \E\frac{\partial}{\partial\bbeta}\bphi(Y_i,\bX_i;\bbeta^*)|| = o_p(1)$. By Slutsky's theorem, $||\bAhat -\bA|| = o_p(1)$. 
Returning to the expansion from the mean value theorem, we have:
\begin{align*}
    n^{1/2}(\bbetahatppi - \bbeta^*) &= \bA n^{-1/2}\sum_{i=1}^n \bphi(Y_i,\bX_i;\bbeta^*)\frac{R_i}{\pi_n} + \bphi(\Yhat_i,\bX_i;\bbeta^*)(1-\frac{R_i}{\pi_n}) \\
    &\quad (\bAhat - \bA) n^{-1/2}\sum_{i=1}^n \bphi(Y_i,\bX_i;\bbeta^*)\frac{R_i}{\pi_n} + \bphi(\Yhat_i,\bX_i;\bbeta^*)(1-\frac{R_i}{\pi_n}) \\
    &= \bA n^{-1/2}\sum_{i=1}^n \bphi(Y_i,\bX_i;\bbeta^*)\frac{R_i}{\pi} + \bphi(\Yhat_i,\bX_i;\bbeta^*)(1-\frac{R_i}{\pi}) \\
    &\quad + \bA n^{-1/2}\sum_{i=1}^n \left\{\bphi(Y_i,\bX_i;\bbeta^*)-\bphi(\Yhat_i,\bX_i;\bbeta^*)\right\}R_i (\pi_n^{-1}-\pi^{-1}) \\
    &\quad + (\bAhat - \bA) n^{-1/2}\sum_{i=1}^n \bphi(Y_i,\bX_i;\bbeta^*)\frac{R_i}{\pi} + \bphi(\Yhat_i,\bX_i;\bbeta^*)(1-\frac{R_i}{\pi}) \\
    &\quad + (\bAhat - \bA) n^{-1/2}\sum_{i=1}^n \left\{\bphi(Y_i,\bX_i;\bbeta^*) - \bphi(\Yhat_i,\bX_i;\bbeta^*)\right\}R_i(\pi_n^{-1}-\pi^{-1}).
\end{align*}
Additionally, note that:
\begin{align*}
    &n^{-1/2}\sum_{i=1}^n\left\{\bphi(Y_i,\bX_i;\bbeta^*)-\bphi(\Yhat_i,\bX_i;\bbeta^*)\right\}R_i (\pi_n^{-1}-\pi^{-1}) \\
    &\quad = n^{-1}\sum_{i=1}^n \left\{\bphi(\Yhat_i,\bX_i;\bbeta^*)-\bphi(Y_i,\bX_i;\bbeta^*)\right\}R_i\pi_n^{-1}\pi^{-1} \left\{n^{-1/2}\sum_{i=1}^n(R_i-\pi)\right\} \\
    &\quad = \E \left\{\bphi(\Yhat_i,\bX_i;\bbeta^*) - \bphi(Y_i,\bX_i;\bbeta^*)\right\}\left\{n^{-1/2}\sum_{i=1}^n \frac{R_i}{\pi}-1\right\} + o_p(1) \\
    &\quad = \E \left\{\bphi(\Yhat_i,\bX_i;\bbeta^*)\right\}\left\{n^{-1/2}\sum_{i=1}^n \frac{R_i}{\pi}-1\right\} + o_p(1),
\end{align*}
where we use the law of large numbers, central limit theorem, and Slutsky's theorem in the second equality and that $\E\bphi(Y_i,\bX_i;\bbeta^*)=\bzero$ in the third. Plugging this into the expansions above and by further applications of the central limit theorem and Slutsky's theorem, we conclude:
\begin{align*}
    &n^{1/2}(\bbetahatppi - \bbeta^*) \\
    &\quad = \bA n^{-1/2}\sum_{i=1}^n \bphi(Y_i,\bX_i;\bbeta^*)\frac{R_i}{\pi} + \left\{\bphi(\Yhat_i,\bX_i;\bbeta^*) - \E\bphi(\Yhat_i,\bX_i;\bbeta^*)\right\}(1-\frac{R_i}{\pi}) + o_p(1).
\end{align*}}
\end{proof}

\blue{\begin{corollary}[Efficiency of $\bbetahatlab$ vs $\bbetahatppi$]\label{coro: betappi}
Let $\bD_{12} =  \E \left\{ \bphi(Y, \bX; \bbeta^*) \bphi(\Yhat, \bX; \bbeta^*)\trans \right\}$ and $\bD_{22} =  Var \bphi(\Yhat, \bX; \bbeta^*)$. The difference in the asymptotic variance between
$\bbetahatlab$ and $\bbetahatppi$ is $\Delta_{\var}( \bbetahatlab, \bbetahatppi) =  (\pi^{-1} - 1) \bA \left(  2  \bD_{12} -  \bD_{22}  \right) \bA\trans$. Consequently, $\bbetahatppi$ is more efficient than $\bbetahatlab$ if and only if $2\bD_{12} - \bD_{22}\succ \bzero$.
\end{corollary}}
\blue{\begin{proof}
Let $\bD_{11}=\E\left\{ \bphi(Y,\bX;\bbeta^*)\bphi(Y,\bX;\bbeta^*)\right\}$. From Proposition 1, the central limit theorem and Slutsky's theorem yields:
\begin{align*}
    n^{1/2}(\bbetahatppi - \bbeta^*) \overset{d}{\to} N(\bzero, \bA\left\{\pi^{-1}\bD_{11} + (\pi^{-1}-1)\bD_{22} + 2(1-\pi^{-1})\bD_{12} \right\}\bA\trans)
\end{align*}
By analogous arguments to those for Proposition 1, it can be shown that $n^{1/2}(\bbetahatlab - \bbeta^*) = n^{-1/2}\bA\sum_{i=1}^n \bphi(Y_i,\bX_i;\bbeta^*)\frac{R_i}{\pi} + o_p(1)$ and thus by the central limit theorem:
\begin{align*}
    n^{1/2}(\bbetahatlab-\bbeta^*) \overset{d}{\to} N(\bzero, \pi^{-1}\bA\bD_{11}\bA\trans).
\end{align*}
The difference in the asymptotic variances is $\Delta_{\var}( \bbetahatlab, \bbetahatppi) = (\pi^{-1} - 1 )  \bA \left(  2  \bD_{12} -  \bD_{22}  \right) \bA\trans$.
\end{proof}}

\subsection{Results for \texttt{CC}}
\blue{\begin{prop}[Asymptotic linearity of $\bbetahatcc$]\label{prop: betacc}
Under the basic assumptions in { Appendix \ref{subse:assump}}$, \bbetahatcc$ is asymptotically linear such that:
\begin{align*}
    n^{1/2}(\bbetahatcc -\bbeta^*) = n^{-1/2}\sum_{i=1}^n {\bf{IF}}^{\cc}(Y_i, \Yhat_i,  \bX_i; \bbeta^*,\bgamma^*) + o_p(1),
\end{align*}
where ${\bf{IF}}^{\cc}(Y_i, \Yhat_i,  \bX_i; \bbeta^*,\bgamma^*) = \bA\left\{\bphi(Y_i,\bX_i;\bbeta^*)\frac{R_i}{\pi} + \bW^{\cc} \bB \bphi(\Yhat_i,\bX_i;\bgamma^*)(1-\frac{R_i}{\pi})\right\}$ and $\bB= \left\{-\E \frac{\partial}{\partial\bgamma\trans} \bphi(\Yhat,\bX;\bgamma^*)\right\}^{-1}$.
\end{prop}}
\blue{\begin{proof}
By analogous arguments to those for Proposition \ref{prop: betappi}, it can be shown that:
\begin{align*}
    &n^{1/2}(\bbetahatlab-\bbeta^*) = n^{-1/2}\bA\sum_{i=1}^n \bphi(Y_i,\bX_i;\bbeta^*)\frac{R_i}{\pi} + o_p(1) \\
    &n^{1/2}(\bgammahatlab-\bgamma^*) = n^{-1/2}\bB\sum_{i=1}^n\bphi(\Yhat_i,\bX_i;\bgamma^*)\frac{R_i}{\pi} + o_p(1)\\
    &n^{1/2}(\bgammahatall-\bgamma^*) = n^{-1/2}\bB\sum_{i=1}^n\bphi(\Yhat_i,\bX_i;\bgamma^*) + o_p(1).
\end{align*}
By application of central limit theorem and Slutsky's theorem:
\begin{align*}
    n^{1/2}(\bbetahatcc-\bbeta^*) &= n^{1/2}(\bbetahatlab-\bbeta^*) -\bA\bW^{\cc} n^{1/2}(\bgammahatlab-\bgamma^* - \bgammahatall + \bgamma^*) \\
    &\quad -(\bAhat\bWhat^{\cc} - \bA\bW^{\cc}) n^{1/2}(\bgammahatlab-\bgamma^* - \bgammahatall + \bgamma^*) \\
    &= n^{-1/2}\bA\sum_{i=1}^n \bphi(Y_i,\bX_i;\bbeta^*)\frac{R_i}{\pi} + \bW^{\cc} \bB \bphi(\Yhat_i,\bX_i;\bgamma^*)(1-\frac{R_i}{\pi}) + o_p(1).
\end{align*}
\end{proof}}

\blue{\begin{corollary}[Efficiency of $\bbetahatlab$ vs $\bbetahatcc$]\label{coro: betacc}
Let $\bC_{12} =  \E \left\{ \bphi(Y, \bX; \bbeta^*) \bphi(\Yhat, \bX; \bgamma^*)\trans \right\}$ and $\bC_{22} =  \E \left\{ \bphi(\Yhat, \bX; \bgamma^*) \bphi(\Yhat, \bX; \bgamma^*)\trans \right\}$. The difference in the asymptotic variance of 
$\bbetahatlab$ and $\bbetahatcc$ is $\Delta_{\var}( \bbetahatlab, \bbetahatcc) =  \bA\bC_{12}\bC_{22}^{-1}\bC_{12}\trans\bA\trans (\pi^{-1}-1)$. Consequently, $\bbetahatcc$ is at least as efficient as $\bbetahatlab$ and is more efficient if and only if $\bC_{12} \neq \bzero$.
\end{corollary}}
\blue{\begin{proof}
First note that $\bW^{\cc}=\bC_{12}\bC_{22}^{-1}\bB^{-1}$. From {Proposition \ref{prop: betacc}}, by the central limit theorem, we have that $n^{1/2}(\bbetahatcc-\bbeta^*) \overset{d}{\to}N(\bzero, Var\{ {\bf{IF}}^{\cc}(Y_i, \Yhat_i,  \bX_i; \bbeta^*,\bgamma^*)\})$, where:
\begin{align*}
    &Var\{ {\bf{IF}}^{\cc}(Y_i, \Yhat_i,  \bX_i; \bbeta^*,\bgamma^*)\} \\
    &\qquad= \pi^{-1}\bA\bD_{11}\bA\trans + \bA\bW^{\cc}\bB\bC_{22}\bB\bW^{\cc\trans}\bA\trans(\pi^{-1}-1) + 2\bA\bC_{12}\bC_{22}^{-1}\bC_{12}\trans\bA\trans (1-\pi^{-1}) \\
    &\qquad= \pi^{-1}\bA\bD_{11}\bA\trans + \bA\bC_{12}\bC_{22}^{-1}\bC_{12}\trans\bA\trans(\pi^{-1}-1) +2\bA\bC_{12}\bC_{22}^{-1}\bC_{12}\bA\trans (1-\pi^{-1}) \\
    &\qquad= \pi^{-1}\bA\bD_{11}\bA\trans +\bA\bC_{12}\bC_{22}^{-1}\bC_{12}\bA\trans (1-\pi^{-1}).
\end{align*}
By analogous arguments to those for Proposition \ref{prop: betappi}, it can be shown that $n^{1/2}(\bbetahatlab - \bbeta^*) = n^{-1/2}\bA\sum_{i=1}^n \bphi(Y_i,\bX_i;\bbeta^*)\frac{R_i}{\pi} + o_p(1)$ and thus $n^{1/2}(\bbetahatlab-\bbeta^*) \overset{d}{\to} N(\bzero, \pi^{-1}\bA\bD_{11}\bA\trans)$. Hence the difference in asymptotic variance is $\Delta_{\var}( \bbetahatlab, \bbetahatcc) =     \bA\bC_{12}\bC_{22}^{-1}\bC_{12}\bA\trans (\pi^{-1}-1)$.
\end{proof}}

\blue{\begin{prop}[$\bbetahatcc$ the most efficient among augmented estimators] \label{prop: betaCC} Let $\bbetatilde^{\cc}(\bVhat)=\bbetahatlab - \bVhat(\bgammahatlab-\bgammahatall)$ be an alternative weighted augmented estimator with weights $\bVhat$ that are consistent for some $\bV^* \in \mathbb{R}^{q\times q}$. Then the difference in asymptotic variance is such that $\Delta_{\var}(\bbetatilde^{\cc}(\bVhat), \bbetahatcc) \succeq \bzero$.
\end{prop}}
\blue{\begin{proof}
First, using similar arguments as those for Proposition \ref{prop: betacc}, $\bbetatilde^{\cc}(\bVhat)$ admits the expansion $n^{1/2}\left\{\bbetatilde^{\cc}(\bVhat)-\bbeta^*\right\} = n^{-1/2}\sum_{i=1}^n {\bf{IF}}(Y_i, \Yhat_i,  \bX_i; \bW^*) + o_p(1)$, where
\begin{align*}
    {\bf{IF}}(Y_i, \Yhat_i,  \bX_i; \bW) = \bA\left\{\bphi(Y_i,\bX_i;\bbeta^*)\frac{R_i}{\pi} + \bW \bB \bphi(\Yhat_i,\bX_i;\bgamma^*)(1-\frac{R_i}{\pi})\right\}
\end{align*}
and $\bW^* = \bA^{-1}\bV^*\bB^{-1}$. The asymptotic variance of $\bbetatilde^{\cc}(\bVhat)$ can be expressed as:
\begin{align*}
    &\var\left\{ {\bf{IF}}(Y_i, \Yhat_i,  \bX_i; \bW^*) \right\} = \var\left\{ {\bf{IF}}(Y_i, \Yhat_i,  \bX_i; \bW^*) -  {\bf{IF}}(Y_i, \Yhat_i,  \bX_i; \bW^\cc) \right\} + \var\left\{{\bf{IF}}(Y_i, \Yhat_i,  \bX_i; \bW^\cc) \right\} \\
    &\qquad + 2 \cov\left\{ {\bf{IF}}(Y_i, \Yhat_i,  \bX_i; \bW^*) -  {\bf{IF}}(Y_i, \Yhat_i,  \bX_i; \bW^\cc),  {\bf{IF}}(Y_i, \Yhat_i,  \bX_i; \bW^\cc) \right\}.
\end{align*}
It thus suffices to show that $\cov \left\{{\bf{IF}}(Y_i, \Yhat_i,  \bX_i; \bW^*)  -  {\bf{IF}}(Y_i, \Yhat_i,  \bX_i; \bW^\cc), {\bf{IF}}(Y_i, \Yhat_i,  \bX_i; \bW^\cc)  \right\} = \bzero.$
To this end, by definition of $\bW^\cc =   \bC_{12} \bC_{22}^{-1}  \bB^{-1}$,
\begin{align*}
&\cov \left\{{\bf{IF}}(Y_i, \Yhat_i,  \bX_i; \bW^*)  -  {\bf{IF}}(Y_i, \Yhat_i,  \bX_i; \bW^\cc), {\bf{IF}}(Y_i, \Yhat_i,  \bX_i; \bW^\cc)  \right\} \\
&= \bA \E\left[ \left\{(\bW^* -\bW^\cc) \bB \bphi(\Yhat_i,\bX_i;\bgamma^*) \left(1-\frac{R_i}{\pi} \right)\right\}  \left\{\bphi(Y_i,\bX_i;\bbeta^*)\frac{R_i}{\pi} + \bW^\cc \bB \bphi(\Yhat_i,\bX_i;\bgamma^*)(1-\frac{R_i}{\pi})\right\}\trans   \right] \bA \\
&= \bA (\bW -\bW^\cc) \bB  \E\left[ \left\{\bphi(\Yhat_i,\bX_i;\bgamma^*) \left[\bphi(Y_i,\bX_i;\bbeta^*) \right]\trans \left(1-\frac{R_i}{\pi} \right) \frac{R_i}{\pi}\right\}  \right] \bA \\
&+ \bA (\bW -\bW^\cc) \bB \E\left[ \left\{ \bphi(\Yhat_i,\bX_i;\bgamma^*) \left[\bphi(\Yhat_i,\bX_i;\bgamma^*) \right]\trans \left(1-\frac{R_i}{\pi} \right)^2\right\}  \right] \bB (\bW^\cc)\trans \bA   \\
&= (\pi^{-1}-1)  \bA(\bW -\bW^\cc) \bB \left\{ \bC_{22} \bB (\bW^\cc)\trans - \bC_{12}\trans \right\}  \bA\\  
&= (\pi^{-1}-1)  \bA(\bW -\bW^\cc) \bB \left( \bC_{12}\trans - \bC_{12}\trans  \right)  \bA \\
&= \bzero.
\end{align*}
\end{proof}}

\subsection{Results for \texttt{PDC}}
\blue{\begin{prop}[Asymptotic linearity of $\bbetahatpdc$] \label{prop: betapdc}
Under the basic assumptions in { Appendix \ref{subse:assump}}, $\bbetahatpdc$ is consistent such that $||\bbetahatpdc -\bbeta^*|| = o_p(1)$. Moreover, $\bbetahatpdc$ is asymptotically linear such that:
\begin{align*}
    n^{1/2}(\bbetahatpdc-\bbeta^*) = n^{-1/2}\sum_{i=1}^n {\bf{IF}}^{\pdc}(Y_i, \Yhat_i,  \bX_i; \bbeta^*) + o_p(1),
\end{align*}
where ${\bf{IF}}^{\pdc}(Y_i, \Yhat_i,  \bX_i; \bbeta^*) = \bA\left[\bphi(Y_i,\bX_i;\bbeta^*)\frac{R_i}{\pi} + \bW^{\pdc} \left\{\bphi(\Yhat_i,\bX_i;\bbeta^*)-\E\bphi(\Yhat_i,\bX_i;\bbeta^*)\right\}(1-\frac{R_i}{\pi})\right]$ and $\bW^{\pdc} = \bD_{12}\bD_{22}^{-1}$.
\end{prop}}
\blue{\begin{proof}
    Let $\bPsi_n^{\pdc}(\bbeta) = n^{-1} \sum_{i = 1}^{n} \bphi(Y_i, \bX_i; \bbeta)\frac{R_i}{\pi_n} + \bWhat^{\pdc}  \bphi(\Yhat_i, \bX_i; \bbeta)  (1-\frac{R_i}{\pi_n})$. Now note that
    \begin{align*}
        &\sup_{\bbeta}||\bPsi_n^{\pdc}(\bbeta) - \bPsi(\bbeta)|| \leq \sup_{\bbeta}||n^{-1}\sum_{i=1}^n\bphi(Y_i,\bX_i;\bbeta)\frac{R_i}{\pi} - \bPsi(\bbeta)|| \\
        &\quad + \sup_{\bbeta}||n^{-1}\sum_{i=1}^n\bphi(Y_i,\bX_i;\bbeta)R_i - \E\left\{ \bphi(Y_i,\bX_i;\bbeta)R_i\right\}|| \cdot |\pi_n^{-1}-\pi^{-1}| \\
        &\quad + \sup_{\bbeta}||\E\left\{ \bphi(Y_i,\bX_i;\bbeta)R_i\right\}|| \cdot |\pi_n^{-1}-\pi^{-1}| \\
        &\quad + ||\bW^{\pdc}|| \cdot \sup_{\bbeta}|| n^{-1}\sum_{i=1}^n \bphi(\Yhat_i,\bX_i;\bbeta)(1-\frac{R_i}{\pi})|| \\
        &\quad + ||\bW^{\pdc}|| \cdot \sup_{\bbeta}||n^{-1}\sum_{i=1}^n\bphi(\Yhat_i,\bX_i;\bbeta)R_i - \E\left\{ \bphi(\Yhat_i,\bX_i;\bbeta)R_i\right\}|| \cdot |\pi^{-1} -\pi_n^{-1}| \\
        &\quad + ||\bW^{\pdc}||\cdot ||\E\left\{ \bphi(\Yhat_i,\bX_i;\bbeta)R_i\right\}|| \cdot ||\pi^{-1}-\pi_n^{-1}|| \\
        &\quad + ||\bWhat^{\pdc}-\bW^{\pdc}|| \cdot || n^{-1}\sum_{i=1}^n \bphi(\Yhat_i,\bX_i;\bbeta)(1-\frac{R_i}{\pi})||\\
        &\quad + ||\bWhat^{\pdc}-\bW^{\pdc}|| \cdot ||n^{-1}\sum_{i=1}^n\bphi(\Yhat_i,\bX_i;\bbeta)R_i - \bW^{\pdc}\E\left\{ \bphi(\Yhat_i,\bX_i;\bbeta)R_i\right\}|| \cdot |\pi^{-1} -\pi_n^{-1}|\\
        &\quad + ||\bWhat^{\pdc}-\bW^{\pdc}|| \cdot ||\E\left\{ \bphi(\Yhat_i,\bX_i;\bbeta)R_i\right\}|| \cdot ||\pi^{-1}-\pi_n^{-1}||.
    \end{align*}
    Using that $\bWhat^{\pdc}$ is consistent such that $||\bWhat^{\pdc}-\bW^{\pdc}|| = o_p(1)$ and arguments analogous to those from {Proposition \ref{prop: betappi}}, it can be shown that the terms on the right-hand side are $o_p(1)$ and thus $\sup_{\bbeta}||\bPsi_n^{\pdc}(\bbeta) - \bPsi(\bbeta)||=o_p(1)$. As $\bPsi(\bbeta)$ has a unique zero by assumption, standard arguments (e.g., Theorem 2.1 in \cite{newey1994large}) verify that $||\bbetahatpdc-\bbeta^*||=o_p(1)$.
    By the mean value theorem, expanding $\bzero = \bPsi_n^{\pdc}(\bbetahatpdc)$ around $\bbeta^*$ yields $n^{1/2}(\bbetahatpdc -\bbeta^*) = \bAhat^{\pdc}n^{1/2}\bPsi_n^{\pdc}(\bbeta^*)$, where $\bAhat^{\pdc} = \left\{ -\frac{\partial}{\partial\bbeta}\bPsi_n^{\pdc}(\bbetatilde^{\pdc})\right\}^{-1}$ and $\bbetatilde^{\pdc}$ is such that $||\bbetatilde^{\pdc}-\bbeta^*|| \leq ||\bbetahatpdc-\bbeta^*||$. By similar arguments to those in Proposition \ref{prop: betappi} and that $||\bWhat^{\pdc}-\bW^{\pdc}|| = o_p(1)$, it can be shown that $||\bAhat^{\pdc}-\bA||=o_p(1)$.  
    Returning to the expansion from the mean value theorem, we have:
    \begin{align*}
        n^{1/2}(\bbetahatpdc-\bbeta^*) &=\bA n^{-1/2}\sum_{i=1}^n \bphi(Y_i,\bX_i;\bbeta^*)\frac{R_i}{\pi_n}+\bW^{\pdc}\bphi(\Yhat_i,\bX_i;\bbeta^*)(1-\frac{R_i}{\pi_n})\\
        &\quad + (\bAhat^{\pdc}-\bA)n^{-1/2}\sum_{i=1}^n \bphi(Y_i,\bX_i;\bbeta^*)\frac{R_i}{\pi_n}+\bW^{\pdc}\bphi(\Yhat_i,\bX_i;\bbeta^*)(1-\frac{R_i}{\pi_n}) \\
        &\quad + \bA(\bWhat^{\pdc}-\bW^{\pdc})n^{-1/2}\sum_{i=1}^n\bphi(\Yhat_i,\bX_i;\bbeta^*)(1-\frac{R_i}{\pi_n}) \\
        &\quad + (\bAhat-\bA)(\bWhat^{\pdc}-\bW^{\pdc})n^{-1/2}\sum_{i=1}^n\bphi(\Yhat_i,\bX_i;\bbeta^*)(1-\frac{R_i}{\pi_n}).
    \end{align*}
    Now we have the expansion:
    \begin{align*}
        &n^{-1/2}\sum_{i=1}^n \bphi(Y_i,\bX_i;\bbeta^*)\frac{R_i}{\pi_n}+\bW^{\pdc}\bphi(\Yhat_i,\bX_i;\bbeta^*)(1-\frac{R_i}{\pi_n}) \\
        &\quad = n^{-1/2}\sum_{i=1}^n \bphi(Y_i,\bX_i;\bbeta^*)\frac{R_i}{\pi}+\bW^{\pdc}\bphi(\Yhat_i,\bX_i;\bbeta^*)(1-\frac{R_i}{\pi}) \\
        &\qquad + n^{-1/2}\sum_{i=1}^n\left\{ \bphi(Y_i,\bX_i;\bbeta^*)-\bW^{\pdc}\bphi(\Yhat_i,\bX_i;\bbeta^*)\right\}R_i(\pi_n^{-1}-\pi^{-1}) \\
        &\quad = n^{-1/2}\sum_{i=1}^n \bphi(Y_i,\bX_i;\bbeta^*)\frac{R_i}{\pi}+\bW^{\pdc}\bphi(\Yhat_i,\bX_i;\bbeta^*)(1-\frac{R_i}{\pi}) \\
        &\qquad - n^{-1}\sum_{i=1}^n\left\{ \bphi(Y_i,\bX_i;\bbeta^*)-\bW^{\pdc}\bphi(\Yhat_i,\bX_i;\bbeta^*)\right\}R_i\pi_n^{-1}\pi^{-1}n^{1/2}(\pi_n-\pi)\\
        &\quad = n^{-1/2}\sum_{i=1}^n \bphi(Y_i,\bX_i;\bbeta^*)\frac{R_i}{\pi}+\bW^{\pdc}\bphi(\Yhat_i,\bX_i;\bbeta^*)(1-\frac{R_i}{\pi}) \\
        &\qquad - n^{-1}\sum_{i=1}^n\left\{ \bphi(Y_i,\bX_i;\bbeta^*)-\bW^{\pdc}\bphi(\Yhat_i,\bX_i;\bbeta^*)\right\}R_i\pi_n^{-1}n^{-1/2}\sum_{i=1}^n(\frac{R_i}{\pi}-1)\\
        &\quad = n^{-1/2}\sum_{i=1}^n \bphi(Y_i,\bX_i;\bbeta^*)\frac{R_i}{\pi}+\bW^{\pdc}\bphi(\Yhat_i,\bX_i;\bbeta^*)(1-\frac{R_i}{\pi}) \\
        &\qquad + \bW^{\pdc} \E \bphi(\Yhat_i,\bX_i;\bbeta^*)n^{-1/2}\sum_{i=1}^n(\frac{R_i}{\pi}-1) + o_p(1)\\
        &\quad = n^{-1/2}\sum_{i=1}^n \bphi(Y_i,\bX_i;\bbeta^*)\frac{R_i}{\pi}+\bW^{\pdc}\left\{\bphi(\Yhat_i,\bX_i;\bbeta^*)-\E \bphi(\Yhat_i,\bX_i;\bbeta^*)\right\}(1-\frac{R_i}{\pi}) + o_p(1),
    \end{align*}
    where the fourth equality applies the law of large numbers, central limit theorem, and Slutsky's theorem. Additionally we have the expansion:
    \begin{align*}
        &n^{-1/2}\sum_{i=1}^n \bphi(\Yhat_i,\bX_i;\bbeta^*)(1-\frac{R_i}{\pi_n}) = n^{-1/2}\sum_{i=1}^n \bphi(\Yhat_i,\bX_i;\bbeta^*)(1-\frac{R_i}{\pi}) - n^{-1/2}\sum_{i=1}^n\bphi(\Yhat_i,\bX_i;\bbeta^*)R_i(\pi_n^{-1}-\pi^{-1}) \\
        &\quad = n^{-1/2}\sum_{i=1}^n \bphi(\Yhat_i,\bX_i;\bbeta^*)(1-\frac{R_i}{\pi}) + n^{-1}\sum_{i=1}^n\bphi(\Yhat_i,\bX_i;\bbeta^*)R_i \pi_n^{-1} n^{-1/2}\sum_{i=1}^n(\frac{R_i}{\pi}-1) \\
        &\quad = n^{-1/2}\sum_{i=1}^n \bphi(\Yhat_i,\bX_i;\bbeta^*)(1-\frac{R_i}{\pi}) + \E\bphi(\Yhat_i,\bX_i;\bbeta^*) n^{-1/2}\sum_{i=1}^n(\frac{R_i}{\pi}-1) + o_p(1).
    \end{align*}
    Plugging in these expansions into the above expansion and by further applications of the central limit theorem and Slutsky's theorem, we conclude:
    \begin{align*}
        &n^{1/2}(\bbetahatpdc-\bbeta^*) \\
        &\quad =\bA n^{-1/2}\sum_{i=1}^n \bphi(Y_i,\bX_i;\bbeta^*)\frac{R_i}{\pi}+\bW^{\pdc}\left\{\bphi(\Yhat_i,\bX_i;\bbeta^*)-\E\bphi(\Yhat_i,\bX_i;\bbeta^*)\right\}(1-\frac{R_i}{\pi}) + o_p(1).
    \end{align*}
\end{proof}}

\blue{\begin{corollary}[Efficiency of $\bbetahatlab$ vs $\bbetahatpdc$]\label{coro: betapdc}
The difference in the asymptotic variance of $\bbetahatlab$ and $\bbetahatpdc$ is $\Delta_{\var}( \bbetahatlab, \bbetahatpdc) =  \bA\bD_{12}\bD_{22}^{-1}\bD_{12}\bA\trans (\pi^{-1}-1)$. Consequently, $\bbetahatpdc$ is more efficient than $\bbetahatlab$ if and only if $\bD_{12} \neq \bzero$.
\end{corollary}}
\blue{\begin{proof}
From {Proposition \ref{prop: betapdc}}, by the central limit theorem, we have that $n^{1/2}(\bbetahatpdc-\bbeta^*) \overset{d}{\to}N(\bzero, Var\{ {\bf{IF}}^{\pdc}(Y_i, \Yhat_i,  \bX_i; \bbeta^*)\})$, where:
\begin{align*}
    &Var\{ {\bf{IF}}^{\pdc}(Y_i, \Yhat_i,  \bX_i; \bbeta^*)\} \\
    &\quad = \pi^{-1}\bA\bD_{11}\bA\trans + \bA\bW^{\pdc}\bD_{22}\bW^{\pdc\trans}\bA\trans(\pi^{-1}-1) + 2\bA\bD_{12}\bW^{\pdc\trans}\bA\trans (1-\pi^{-1}) \\
    &\quad = \pi^{-1}\bA\bD_{11}\bA\trans + \bA\bD_{12}\bD_{22}^{-1}\bD_{12}\trans\bA\trans(\pi^{-1}-1) +2\bA\bD_{12}\bD_{22}^{-1}\bD_{12}\trans\bA\trans (1-\pi^{-1}) \\
    &\quad = \pi^{-1}\bA\bD_{11}\bA\trans  +\bA\bD_{12}\bD_{22}^{-1}\bD_{12}\trans\bA\trans (1-\pi^{-1}).
\end{align*}
By analogous arguments to those for Proposition 1, it can be shown that $n^{1/2}(\bbetahatlab - \bbeta^*) = n^{-1/2}\bA\sum_{i=1}^n \bphi(Y_i,\bX_i;\bbeta^*)\frac{R_i}{\pi} + o_p(1)$ and thus $n^{1/2}(\bbetahatlab-\bbeta^*) \overset{d}{\to} N(\bzero, \pi^{-1}\bA\bD_{11}\bA\trans)$. Hence the difference in asymptotic variance is $\Delta_{\var}( \bbetahatlab, \bbetahatpdc) =     \bA\bD_{12}\bD_{22}^{-1}\bD_{12}\trans\bA\trans (\pi^{-1}-1)$.
\end{proof}}

\blue{\begin{prop}[Efficiency of $\bbetahatcc$ vs.\ $\bbetahatpdc$ for GLMs] \label{prop: betacc-betapdc-glm} 
Suppose that we seek to estimate regression parameters from a GLM with canonical link such that $\bphi(y,\bx;\bbeta) = \bx\left\{y-g(\bx\trans\bbeta)\right\}$ with $g(\cdot)$ being the inverse link function. A sufficient condition for $\bbetahatcc$ to be at least as efficient as $\bbetahatpdc$ such that $\Delta_{\var}( \bbetahatpdc, \bbetahatcc)\succeq \bzero$ is for the conditional means of both $Y$ and $\Yhat$ to be correctly specified such that:
\begin{align*}
    \E(Y \mid \bX) = g(\bX\trans\bbeta^*) \quad \mbox{and} \quad \E(\Yhat \mid \bX) = g(\bX\trans\bgamma^*).
\end{align*}
\end{prop}}
\blue{\begin{proof}
From Corollaries \ref{coro: betacc} and \ref{coro: betapdc}, $\bbetahatcc$ is more efficient than $\bbetahatpdc$ if and only if $\bA\bC_{12}\bC_{22}^{-1}\bC_{12}\bA(\pi^{-1}-1)  - \bA\bD_{12}\bD_{22}^{-1}\bD_{12}\bA(\pi^{-1}-1)\succ \bzero$. This condition is equivalent to $\bC_{12}\bC_{22}^{-1}\bC_{121} - \bD_{12}\bD_{22}^{-1}\bD_{12}\succ \bzero$.
Now, when $\bphi(y,\bx;\bbeta) = \bx\left\{y-g(\bx\trans\bbeta)\right\}$ and $\E(Y \mid \bX) = g(\bX\trans\bbeta^*)$,
\begin{align*}
    \bD_{12} &= \E \left[ \bX\bX\trans \left\{ Y - g(\bX\trans\bbeta^*) \right\}  \left\{ \Yhat - g(\bX\trans\bbeta^*) \right\} \right] \\
    &= \E \left[  \bX \bX\trans \left[ Y - g(\bX\trans\bbeta^*) \right]  \Yhat \right] \\
    &= \bC_{12}.
\end{align*}
It thus suffices to show that $\bD_{22}-\bC_{22} \succeq \bzero $. But note that when $\E(\Yhat \mid \bX) = g(\bX\trans\bgamma^*)$, $\bC_{22} =  \var \left[\bX \left\{\Yhat - g(\bX\trans\bgamma^*) \right\} \right] = \E \left\{  \var\left(\bX \Yhat  \mid \bX \right)  \right\}$,
while:
\begin{align*}
\bD_{22} &=  \var \left[\bX \left\{\Yhat - g(\bX\trans\beta^*) \right\} \right] \\
&= \E \left\{  \var\left(\bX \Yhat  \mid \bX \right)  \right\} + \var \left\{  \E\left[\bX \left\{\Yhat - g(\bX\trans\beta^*) \right\} \mid \bX \right]  \right\} \\
&= \bC_{22} + \var\left[\bX \left\{g(\bX\trans\bgamma^*) - g(\bX\trans\beta^*) \right\} \right].
\end{align*}
Hence $\bD_{22}-\bC_{22} \succeq \bzero$.
\end{proof}}

\section{Empirical studies}
\subsection{Linear regression}
\blue{Here we derive the expressions for the difference in the asymptotic variance of each PB inference method relative to the classical estimator.}

\blue{\begin{suppprop}[Asymptotic Variance of the PB Inference Methods.]
Following the simulation set-up in Section 6 of the main text, consider
$$Y = \bX_1 \trans \bbeta_1  + \bX_2 \trans \bbeta_2 + \epsilon \quad \mbox{and} \quad \Yhat = \bX_1 \trans \btheta_1  + \bX_2 \trans \btheta_2 +   \bX_3 \trans \btheta_3 $$
where $\epsilon \sim N(0, \sigma^2_{\epsilon})$, $\epsilon \perp (\bX_1, \bX_2, \bX_3)$, and $\bX_1$, $\bX_2$, and $\bX_3$ are independent $d$-dimensional mean-zero normal vectors with variance $\Sigma_{\bX_j} = \sigma_j^2 \bI_d$ for $j = 1, 2, 3$. Let $r = \sigma_2^2 / \sigma_1^2$ and $u = r^{-1}   \| \btheta_1 - \bbeta_1 \|^2  +  \| \btheta_2 \|^2 +   (\sigma_3^2/\sigma_2^2)  \| \btheta_3 \|^2 $. Then, for $m \in \{\texttt{PPI}, \texttt{PDC}, \texttt{CC} \}$
$$\Delta_{\var}( \bbetahatlab_1, \bbetahat^m_1) =  \left( \frac{1}{\pi}  - 1 \right) r \blambda^m$$
where 
$$\blambda^\ppi = \left (\|\bbeta_2\|^2  - \|\btheta_2 - \bbeta_2\|^2 - r^{-1}  \| \btheta_1 - \bbeta_1 \|^2 -  (\sigma_3^2/\sigma_2^2)  \| \btheta_3 \|^2   \right) \bI_d - r^{-1} (\btheta_1 - \bbeta_1)(\btheta_1 - \bbeta_1)\trans, $$
$$\blambda^\pdc = \frac{\dfrac{1}{4} \left( \| \bbeta_2 \|^2 +  \| \btheta_2 \|^2  - \| \btheta_2 - \bbeta_2 \|^2 \right)^2}{r^{-1}   \| \btheta_1 - \bbeta_1 \|^2  +  \| \btheta_2 \|^2 +   (\sigma_3^2/\sigma_2^2)  \| \btheta_3 \|^2} \bI_d - \frac{r^{-1}(\btheta_1 - \bbeta_1)(\btheta_1 - \bbeta_1)\trans}{u^2 + u r^{-1}   \| \btheta_1 - \bbeta_1 \|^2} \mbox{ and } $$
$$ \blambda^\cc = \left(  \dfrac{  \dfrac{1}{4} \left( \| \bbeta_2 \|^2 +  \| \btheta_2 \|^2  - \| \btheta_2 - \bbeta_2 \|^2 \right)^2 }{   \| \btheta_2 \|^2 + (\sigma_3^2 / \sigma_2^2)  \| \btheta_3 \|^2}  \right) \bI_d.$$
\end{suppprop}\label{prop: sim-study}}
\blue{\begin{proof}
Noting that $\bgamma_1^* \coloneq \E( \bX_1 \bX_1 \trans )^{-1} \E(\bX_1 \Yhat) = \btheta_1 $, we first consider the \texttt{CC} estimator and apply Corollary \ref{coro: betacc}. Specifically, $\bC_{22} = (\sigma_1^2 \bm{I}_d) \sigma_{\Yhat}^2$ and $\bC_{12} = (\sigma_1^2 \bm{I}_d)\sigma_{Y, \Yhat}$ where 
\begin{align*}
     \sigma_{\Yhat}^2 &=  \E[ (\bX_2 \trans \btheta_2 + \bX_3 \trans \btheta_3)^2] =  \sigma_2^2 \bI_d  \| \btheta_2 \|^2 + \sigma_3^2 \bI_d  \| \btheta_3 \|^2 
\end{align*}
and
\begin{align*}     
     \sigma_{Y, \Yhat} =   \E[(\bX_2 \trans \bbeta_2  + \epsilon)(\bX_2 \trans \btheta_2 + \bX_3 \trans \btheta_3)] =  \sigma_2^2 \bI_d  (\bbeta_2\trans   \btheta_2). 
\end{align*}  
Letting $r = \sigma_2^2 / \sigma_1^2$, it then follows that $\Delta_{\var}( \bbetahatlab_1, \bbetahatcc_1) = r \left( \frac{1}{\pi}  - 1 \right)  \blambda^\cc$ where 
\begin{align*}
 \blambda^\cc &= r^{-1} \sigma_1^{-4} \bC_{12} \bC_{22}^{-1} \bC_{12} = r^{-1}  \sigma_1^{-2} \left( \frac{\sigma_{Y, \Yhat}^2}{\sigma_{\Yhat}^2} \right) \bI_d
 = r^{-1}  \sigma_1^{-2} \left( \frac{\left(\sigma_2^2 \bI_d  (\bbeta_2\trans   \btheta_2) \right)^2}{\sigma_2^2  \| \btheta_2 \|^2 + \sigma_3^2   \| \btheta_3 \|^2 } \right) \bI_d \\
 &= \left(  \dfrac{  \dfrac{1}{4} \left( \| \bbeta_2 \|^2 +  \| \btheta_2 \|^2  - \| \btheta_2 - \bbeta_2 \|^2 \right)^2 }{   \| \btheta_2 \|^2 + (\sigma_3^2 / \sigma_2^2)  \| \btheta_3 \|^2}  \right) \bI_d.
\end{align*}
For the \texttt{PPI} and \texttt{PDC} estimators, we have that
\begin{align*}
 \bD_{12} &= \E\left[ \bX_1 \bX_1\trans  (Y - \bX_1 \bbeta_1) (\Yhat - \bX_1\trans \bbeta_1)  \right]
 = \bC_{12} + \E \left[ \bX_1 \bX_1\trans  (Y - \bX_1\trans \bbeta_1) (\bX_1\trans (\btheta_1 - \bbeta_1))   \right] = \sigma_1^2 \sigma_2^2 \bI_d  (\bbeta_2\trans   \btheta_2).
\end{align*}
Additionally,
\begin{align*}
\bD_{22} &= \var \left[ \bX_1 (\Yhat - \bX_1\trans \bbeta_1)  \right]  =  \var \left[ \bX_1 (\Yhat - \bX_1\trans \btheta_1)  \right] +  \var \left[ \bX_1 \bX_1\trans (\bbeta_1 - \btheta_1)  \right] \\
&=  \bC_{22} + \E( \bX_1 \bX_1 \trans (\btheta_1 - \bbeta_1)(\btheta_1 - \bbeta_1)\trans \bX_1 \bX_1 \trans) - \sigma_1^4 (\btheta_1 - \bbeta_1) (\btheta_1 - \bbeta_1) \trans   
\end{align*}
To compute $ \E( \bX_1 \bX_1 \trans (\btheta_1 - \bbeta_1)(\btheta_1 - \bbeta_1)\trans \bX_1 \bX_1 \trans)$, we consider its $(i,j)^{th}$ element:
\begin{align*}
& \sum_{k, l} (\theta_{1,k} - \beta_{1,k})  (\theta_{1,l} - \beta_{1,l}) \E \left( X_{1,i} X_{1,j} X_{1,k} X_{1,l} \right)\\
& \sigma_1^4  \sum_{k, l} (\theta_{1,k} - \beta_{1,k})  (\theta_{1,l} - \beta_{1,l}) \left[ I(i=j) I(k= l) + I(i=k) I(j= l)   +   I(i=l) I(j= k) \right]   \\
& = \sigma_1^4 \left[ \| \btheta_1 - \bbeta_1 \|^2 I(i =j) + 2 (\theta_{1,i} - \beta_{1,i})  (\theta_{1,j} - \beta_{1,j})  \right].
\end{align*}
Putting this together,
\begin{align*}
    \bD_{22} =  \sigma_1^2 \left[ \bI_d \left( \sigma_1^2   \| \btheta_1 - \bbeta_1 \|^2  +  \sigma_2^2   \| \btheta_2 \|^2 +   \sigma_3^2   \| \btheta_3 \|^2 \right)  + \sigma_1^2  (\btheta_1 - \bbeta_1)(\btheta_1 - \bbeta_1)\trans  \right]
\end{align*}
For the \texttt{PPI} estimator, it follows from Corollary \ref{coro: betappi} that $\Delta_{\var}( \bbetahatlab_1, \bbetahatppi_1) = \left( \frac{1}{\pi}  - 1 \right) r \blambda^\ppi$
 where
\begin{align*}
\blambda^\ppi &= r^{-1} \sigma_1^{-4} \left(2 \bD_{12} - \bD_{22} \right)\\
&= r^{-1} \sigma_1^{-2}  \left[ \sigma_2^2 (\|\bbeta_2\|^2 + \|\btheta_2\|^2  - \|\btheta_2 - \bbeta_2\|^2 )\right] \bI_d \\
& - r^{-1} \sigma_1^{-2}  \left[ \sigma_1^2   \| \btheta_1 - \bbeta_1 \|^2  +  \sigma_2^2   \| \btheta_2 \|^2 +   \sigma_3^2   \| \btheta_3 \|^2  )\right] \bI_d - r^{-1} \sigma_1^{-2} \left[\sigma_1^2  (\btheta_1 - \bbeta_1)(\btheta_1 - \bbeta_1)\trans  \right]\\
&= \left (\|\bbeta_2\|^2  - \|\btheta_2 - \bbeta_2\|^2 - r^{-1}  \| \btheta_1 - \bbeta_1 \|^2 -  (\sigma_3^2/\sigma_2^2)  \| \btheta_3 \|^2   \right) \bI_d - r^{-1} (\btheta_1 - \bbeta_1)(\btheta_1 - \bbeta_1)\trans 
\end{align*}
For the \texttt{PDC} estimator, we let $u = r^{-1}   \| \btheta_1 - \bbeta_1 \|^2  +  \| \btheta_2 \|^2 +   (\sigma_3^2/\sigma_2^2)  \| \btheta_3 \|^2 $ and apply the Sherman-Morrison formula to find that 
\begin{align*}
\left(\sigma_2^{-2} \bD_{22} \right)^{-1} = \sigma_1^{-2} \left[ u \bI_d + r^{-1} (\btheta_1 - \bbeta_1)(\btheta_1 - \bbeta_1) \right]^{-1} &=   \sigma_1^{-2} \left(\frac{1}{u} \bI_d - \frac{r^{-1} (\btheta_1 - \bbeta_1)(\btheta_1 - \bbeta_1)\trans}{u^2 + u r^{-1}   \| \btheta_1 - \bbeta_1 \|^2 }\right)
\end{align*}
It then follows from Corollary \ref{coro: betapdc} that
$\Delta_{\var}( \bbetahatlab_1, \bbetahatpdc_1) = \left( \frac{1}{\pi}  - 1 \right) r \blambda^\pdc$
 where
\begin{align*}
\blambda^\pdc &= r^{-1} \sigma_1^{-4} \bD_{12} \bD_{22}^{-1} \bD_{12}  \\
&=  \dfrac{\sigma_1^2 }{4} \left( \| \bbeta_2 \|^2 +  \| \btheta_2 \|^2  - \| \btheta_2 - \bbeta_2 \|^2 \right)^2   (\sigma_2^{-2} \bD_{22})^{-1} \\
&= \frac{\dfrac{1}{4} \left( \| \bbeta_2 \|^2 +  \| \btheta_2 \|^2  - \| \btheta_2 - \bbeta_2 \|^2 \right)^2}{r^{-1}   \| \btheta_1 - \bbeta_1 \|^2  +  \| \btheta_2 \|^2 +   (\sigma_3^2/\sigma_2^2)  \| \btheta_3 \|^2} \bI_d - \frac{r^{-1}(\btheta_1 - \bbeta_1)(\btheta_1 - \bbeta_1)\trans}{u^2 + u r^{-1}   \| \btheta_1 - \bbeta_1 \|^2}.
\end{align*}
\end{proof}}

\subsection{Poisson regression}
\subsubsection{Data generating process}
\blue{For Poisson regression, we follow the same general set-up used for the linear regression simulation study. Specifically, we generated
$$Y \sim \text{Poisson} \left[ \exp \left(\bX_1 \trans \bbeta_1  + \bX_2 \trans \bbeta_2 \right) \right]  \quad \mbox{and} \quad \Yhat = \exp\left(\bX_1 \trans \btheta_1  + \bX_2 \trans \btheta_2 +  \bX_3 \trans \btheta_3 \right) $$
where $\bX_1$, $\bX_2$, and $\bX_3$ are independent $d$-dimensional mean-zero normal vectors with variance $\sigma^2_{\bX_j}\bI_d$ for $j = 1, 2, 3$. We again focus on conducting inference on $\bbeta_1$ and consider the same set of ideal and distributional shift settings described in Table 1 of the main text.} 

\blue{Throughout, we set $d = 2$, $n = 10,000$, $\pi = 0.1$, and generated $\bbeta_1$ and $\bbeta_2$ uniformly from the unit sphere $\mathbb{S}^{d-1}$ in each replicate. In the two ideal sub-settings, we let $\sigma^2_{\bX_1} + \sigma^2_{\bX_2} = 0.8$ and assessed performance across the variance ratio $\sigma^2_{\bX_2}/ \sigma^2_{\bX_1}  \in [0.5,  1.5]$. In the distributional shift scenarios, we let $\sigma^2_{\bX_1} = \sigma^2_{\bX_2} = 0.8$. For sub-setting 1, we generated $\btheta_3$ uniformly from the unit sphere $\mathbb{S}^{d-1}$ in each replicate and assessed performance across $\sigma^2_{\bX_3} / \sigma^2_{\bX_2} \in [0.1, 4]$. For sub-settings 2-4, we generated $\dfrac{\bbeta_j - \btheta_j}{\| \bbeta_j - \btheta_j \|}$ uniformly from the unit sphere $\mathbb{S}^{d-1}$ in each replicate and assessed performance across $\| \bbeta_j - \btheta_j \| \in [0, 2]$. We present the width and coverage probability of the 95\% confidence intervals for the first component of $\bbeta_{1}^*$ averaged over 1000 simulation replicates.}

\subsubsection{Results}
\begin{figure}[H]
    \centering
    \begin{subfigure}[a]{\textwidth}
        \centering
        \includegraphics[width=\textwidth]{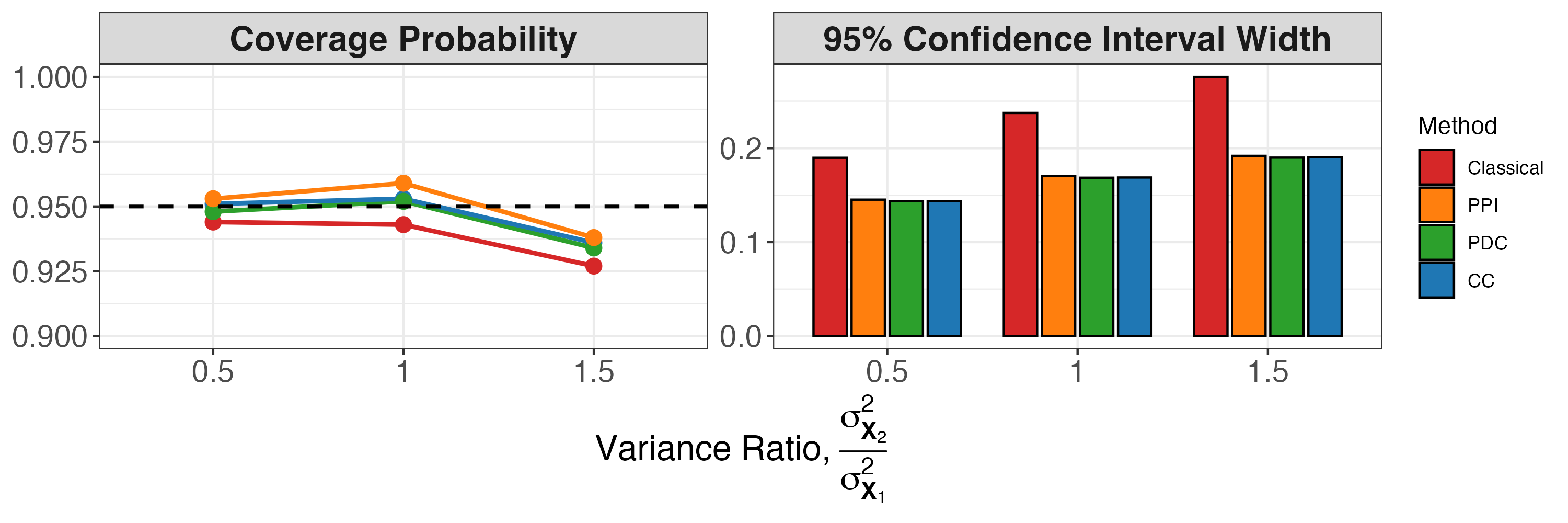}
        \caption{}
          \label{fig: fo-results}
    \end{subfigure}

    \begin{subfigure}[b]{\textwidth}
        \centering
        \includegraphics[width=\textwidth]{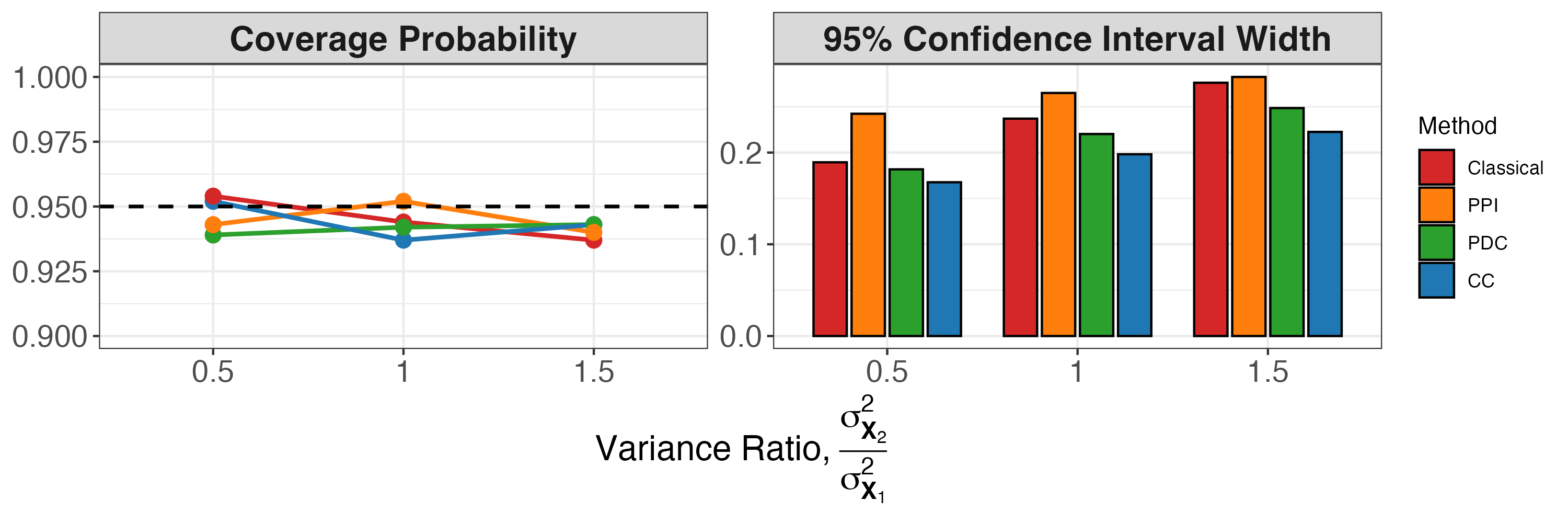}
        \caption{}
          \label{fig: po-results}
    \end{subfigure}

    \begin{subfigure}[c]{\textwidth}
        \centering
        \includegraphics[width=\textwidth]{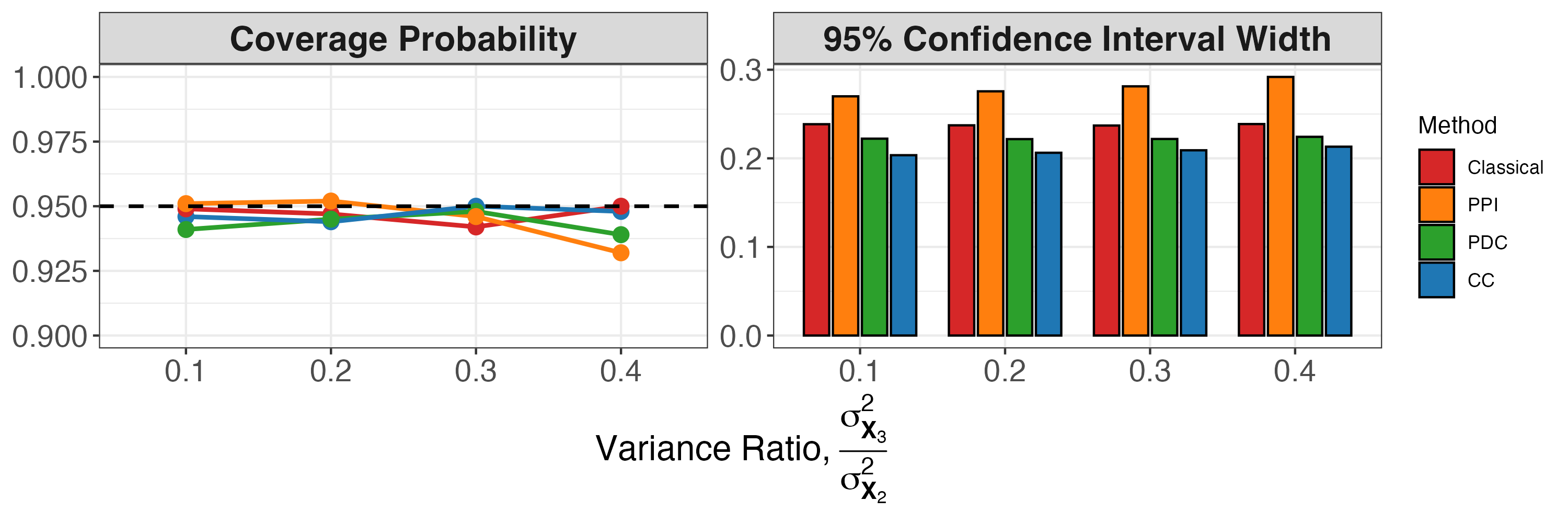}
        \caption{}
          \label{fig: np-results}
    \end{subfigure}

    \caption{Coverage probability and width of the 95\% confidence intervals for the ideal simulation scenario with (a) $\Yhat = \exp \left(\bX_1\trans \bbeta_1 + \bX_2\trans \bbeta_2\right)$ or (b) $\Yhat =\exp\left(\bX_2\trans \bbeta_2\right)$ and the distributional shift scenario with (c) $\Yhat = \exp\left(\bX_2\trans \bbeta_2 + \bX_3\trans \btheta_3\right)$.}
    \label{fig: results-1}
\end{figure}

\begin{figure}[H]
    \centering
    \begin{subfigure}[b]{\textwidth}
        \centering
        \includegraphics[width=\textwidth]{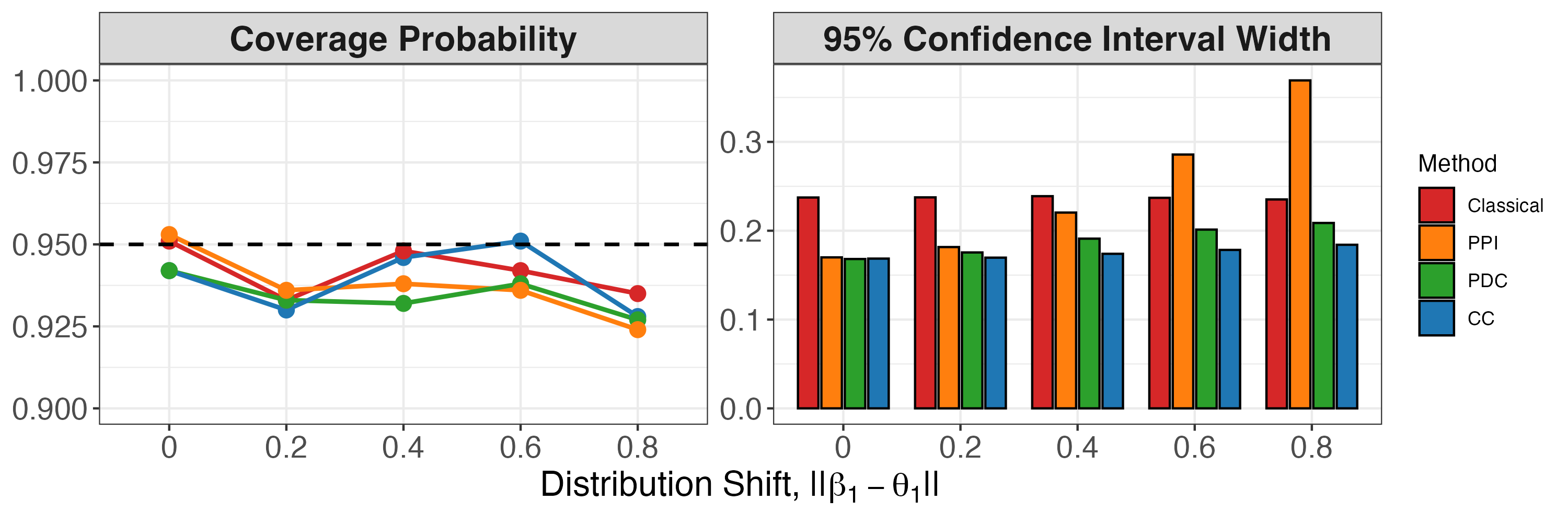}
        \caption{}
          \label{fig: x1s-results}
    \end{subfigure}

    \begin{subfigure}[b]{\textwidth}
        \centering
        \includegraphics[width=\textwidth]{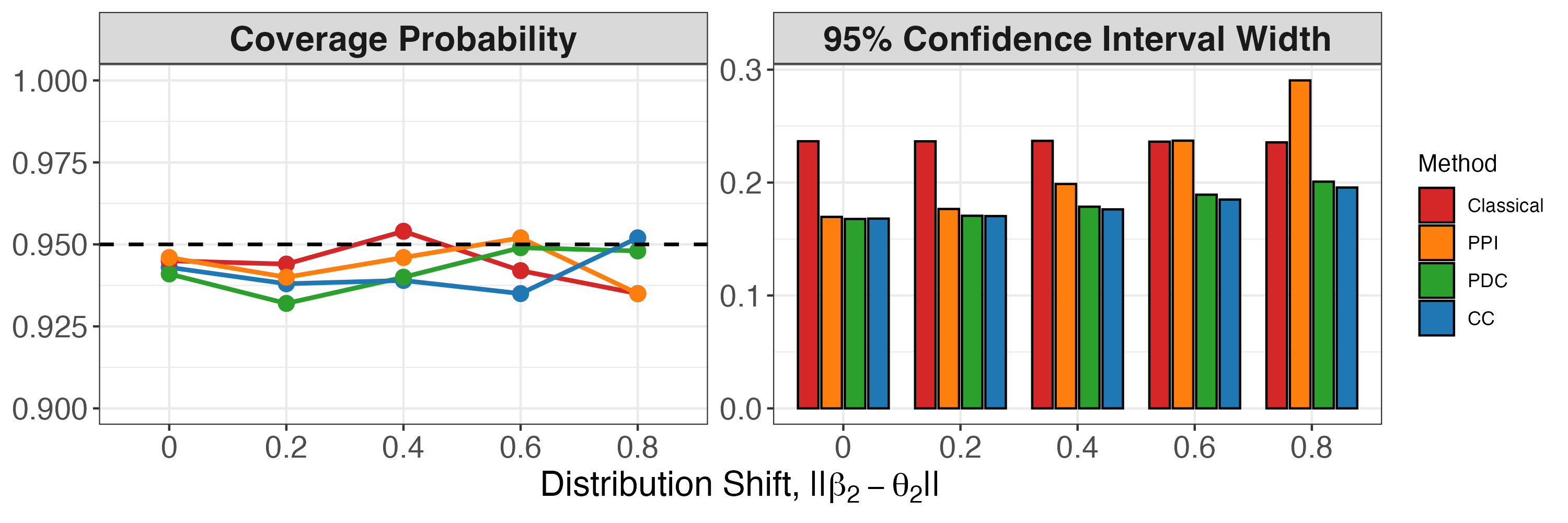}
        \caption{}
          \label{fig: x2s-results}
    \end{subfigure}

     \begin{subfigure}[a]{\textwidth}
        \centering
        \includegraphics[width=\textwidth]{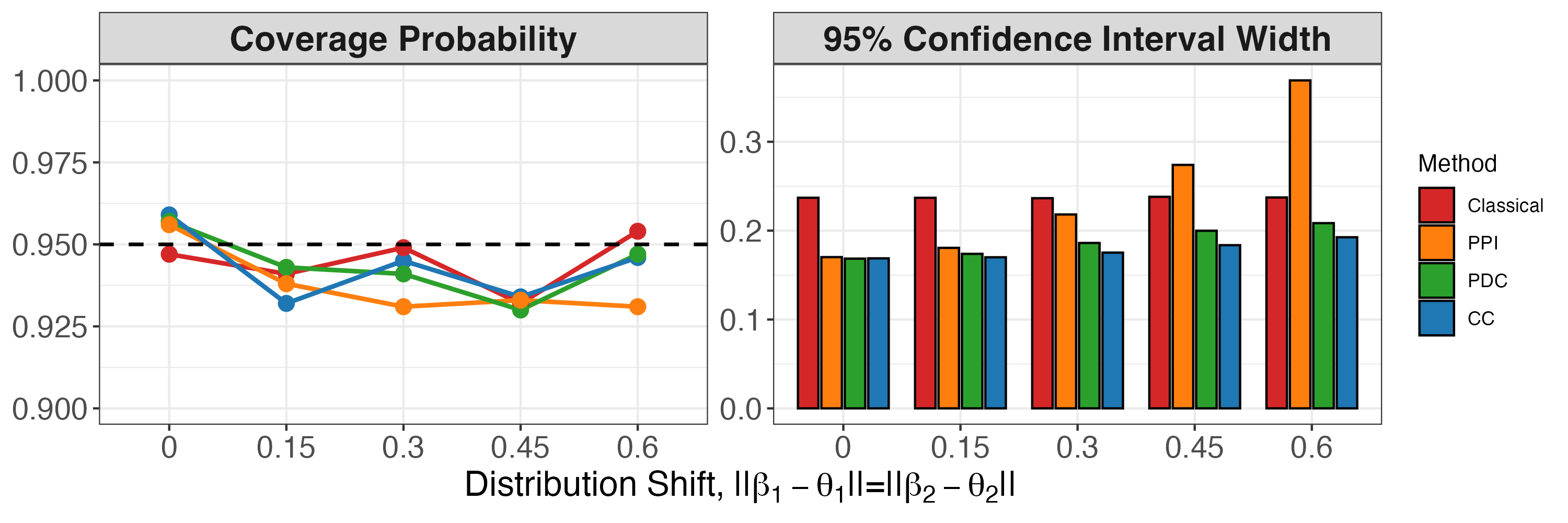}
        \caption{}
          \label{fig: x1x2s-results}
    \end{subfigure}
    
    \caption{Coverage probability and width of the 95\% confidence intervals for the distributional shift simulation scenario with (a) $\Yhat = \exp\left(\bX_1\trans \btheta_1 + \bX_2\trans \bbeta_2\right)$, (b) $\Yhat = \exp\left(\bX_1\trans \bbeta_1 + \bX_2\trans \btheta_2\right)$ or (c) $\Yhat = \exp\left(\bX_1\trans \btheta_1 + \bX_2\trans \btheta_2\right)$}
    \label{fig: dist-shift-results-1}
\end{figure}

\blue{The general patterns for all settings mimic those for linear regression. The results for the ideal sub-settings are presented in Figures \ref{fig: fo-results} and \ref{fig: po-results} while those for the distributional shift sub-settings are presented in Figures \ref{fig: np-results} and \ref{fig: x1s-results}-\ref{fig: x1x2s-results}.}

\newpage

\newpage

\vskip 0.2in

\bibliographystyle{amawiley}
\bibliography{sampleama}

%% file: GrandMacros.tex
\newcommand{\bZ}{{\bf{Z}}}
\newcommand{\bX}{{\bf{X}}}

\newcommand{\bI}{{\bf{I}}}
\newcommand{\bx}{{\bf{x}}}
\newcommand{\bSigma}{\boldsymbol{\Sigma}}

\newcommand{\fhat}{\hat{f}}
\newcommand{\Yhat}{\widehat{Y}}

\newcommand{\bAhat}{\widehat{\bA}}
\newcommand{\bVhat}{\widehat{\bV}}
\newcommand\bWhat{\widehat{\bW}}

\newcommand{\var}{\text{Var}}

\newcommand{\E}{\mathbb{E}}
\renewcommand{\P}{\mathbb{P}}

\newcommand{\bC}{{\bf{C}}}

\newcommand{\bbeta}{\bm{\beta}}
\newcommand{\btheta}{\bm{\theta}}
\newcommand{\bgamma}{\bm{\gamma}}
\newcommand{\bomega}{\bm{\omega}}
\newcommand{\bphi}{\bm{\phi}}

\newcommand{\bbetahat}{\hat\bbeta}
\newcommand{\bbetatilde}{\widetilde{\bbeta}}
\newcommand{\bgammahat}{\hat\bgamma}
\newcommand{\bW}{{\bf{W}}}
\newcommand{\bV}{{\bf{V}}}
\newcommand{\bD}{{\bf{D}}}

\newcommand{\bzero}{\boldsymbol 0}

\def\transpose{{\sf \scriptscriptstyle{T}}}
\def\trans{^{\transpose}}

\newcommand{\ppi}{\text{\tiny{PPI}}}

\newcommand{\cc}{\text{\tiny{CC}}}
\newcommand{\pdc}{\text{\tiny{PDC}}}

\newcommand{\opt}{\text{\tiny{opt}}}
\newcommand{\sur}{\text{\tiny{SUR}}}

\newcommand{\pop}{\text{\tiny{POP}}}

\newcommand{\bbetahatlab}{\bbetahat^{\text{\tiny{lab}}}}

\newcommand{\bbetahatppi}{\bbetahat^{\text{\tiny{PPI}}}}
\newcommand{\bbetahatpdc}{\bbetahat^{\text{\tiny{PDC}}}}

\newcommand{\bbetahatcc}{\bbetahat^{\cc}}
\newcommand{\bbetahatsur}{\bbetahat^{\sur}}

\newcommand{\bbetahatpop}{\bbetahat^{\pop}}

\newcommand{\bgammahatlab}{\bgammahat^{\text{\tiny{lab}}}}
\newcommand{\bgammahatunlab}{\bgammahat^{\text{\tiny{unlab}}}}
\newcommand{\bgammahatall}{\bgammahat^{\text{\tiny{all}}}}

\newcommand{\nlab}{n_{\text{\tiny{lab}}}}
\newcommand{\nunlab}{n_{\text{\tiny{unlab}}}}

\newcommand{\bA }{ {\bf{A}}}
\newcommand{\bB }{ {\bf{B}}}